\documentclass{article}

\usepackage[preprint]{neurips_2026}

\usepackage[utf8]{inputenc}
\usepackage[T1]{fontenc}
\usepackage{hyperref}
\usepackage{url}
\usepackage{booktabs}
\usepackage{amsfonts}
\usepackage{nicefrac}
\usepackage{microtype}
\usepackage{xcolor}

\usepackage{graphicx}
\usepackage{subcaption}
\usepackage{makecell}
\usepackage{amsmath}
\usepackage{amssymb}
\usepackage{mathtools}
\usepackage{amsthm}
\usepackage[capitalize,noabbrev]{cleveref}
\usepackage[noupquote]{inconsolata}
\usepackage{textcomp}
\usepackage{upquote}
\usepackage{soul}

\usepackage{color}
\usepackage{multirow}
\usepackage{multicol}
\usepackage{enumitem,kantlipsum}
\usepackage[normalem]{ulem}
\usepackage{colortbl}
\usepackage{threeparttable}
\usepackage{tcolorbox}
\usepackage{comment}
\usepackage{array}
\usepackage{lscape}
\useunder{\uline}{\ul}{}
\usepackage{fancybox}
\usepackage{tikz}
\usepackage{listings}
\usepackage{pgfplots}
\pgfplotsset{compat=1.18}

\definecolor{nxOrange}{RGB}{255,127,14}
\definecolor{nxBlue}{RGB}{93,165,218}

\usepackage{pdflscape}
\usepackage{siunitx}
\usepackage{pgfplotstable}

\usepackage{wrapfig}
\definecolor{lightblue}{rgb}{.50,.95,1}
\definecolor{tri}{rgb}{.25,.88,.82}
\definecolor{lilac}{rgb}{0.85,0.64,0.85}

\newcommand{\emqa}{\textbf{\emph{EverydayMMQA}}}
\newcommand{\oasis}{\textbf{\emph{OASIS}}}

\definecolor{lilac}{RGB}{230, 200, 255}

\usepackage{verbatim}
\usepackage{setspace}

\definecolor{codegray}{gray}{0.9}
\definecolor{dkgreen}{rgb}{0,0.6,0}
\definecolor{gray}{rgb}{0.5,0.5,0.5}

\lstdefinestyle{pythonstyle}{
  frame=tb,
  language=python,
  aboveskip=3mm,
  belowskip=3mm,
  showstringspaces=false,
  columns=flexible,
  basicstyle={\small\ttfamily},
  numbers=none,
  numberstyle=\tiny\color{gray},
  keywordstyle=\color{dkgreen},
  commentstyle=\color{dkgreen},
  breaklines=true,
  breakatwhitespace=true,
  tabsize=3,
  escapeinside={`}{`},
  breakindent=0pt,
  otherkeywords={these}
}

\usepackage{arabtex}
\usepackage{utf8}
\setcode{utf8}

\usepackage{textcomp}

\usepackage{wrapfig}
\usepackage{subcaption}
\usepackage{pgfplots}
\pgfplotsset{compat=1.18}

\theoremstyle{plain}

\theoremstyle{definition}

\theoremstyle{remark}

\usepackage[textsize=tiny]{todonotes}

\usepackage{titletoc}

\title{OASIS: A Multilingual and Multimodal Dataset for Culturally Grounded Spoken Visual QA}

\author{Firoj Alam,
    Ali Ezzat Shahroor,
    Md. Arid Hasan\thanks{~~The contribution was made while the author was a contributor at the Qatar Computing Research Institute.},
    Zien Sheikh Ali,
    Hunzalah Hassan Bhatti,\\
    {\bf
    Mohamed Bayan Kmainasi,
    Shammur Absar Chowdhury,
    Basel Mousi,
    Fahim Dalvi,} \\
    {\bf
    Nadir Durrani,
    Natasa Milic-Frayling
    } \\
    Qatar Computing Research Institute, Qatar, \\
    \{fialam, shchowdhury, ndurrani, nmilicfrayling\}@hbku.edu.qa
}

\begin{document}

\maketitle

\begin{abstract}
Large-scale multimodal models achieve strong results on tasks like Visual Question Answering (VQA), but they are often limited when queries require \textit{cultural and visual information}, everyday knowledge, particularly in low-resource and underrepresented languages. We introduce \oasis{}, a large-scale culturally grounded multimodal QA dataset covering images, text, and speech. \oasis{} is built with \emqa{}, a scalable semi-automatic framework for creating localized spoken and visual QA resources, supported by multi-stage human-in-the-loop validation. \oasis{} contains approximately \textit{0.92M real images} and \textit{14.8M QA pairs}, including \textit{3.7M spoken questions}, with \textit{383 hours of human-recorded speech}, and \textit{20K hours of voice-cloned speech}, from 42 speakers. It supports four input settings: text-only, speech-only, text+image, and speech+image. The dataset focuses on English and Arabic varieties across 18 countries, covering Modern Standard Arabic (MSA) as well as dialectal Arabic. It is designed to evaluate models beyond object recognition, targeting pragmatic, commonsense, and culturally grounded reasoning in real-world scenarios. We benchmark four closed-source models, three open-source models, and one fine-tuned model on \oasis{}. The \textit{framework} and \textit{dataset} will be made publicly available to the community.\footnote{\href{https://huggingface.co/datasets/QCRI/OASIS}{https://huggingface.co/datasets/QCRI/OASIS}}

\end{quote}
\end{abstract}
\section{Introduction}
\label{sec:introduction}

\begin{figure*}
\centering
\includegraphics[width=0.96\linewidth]{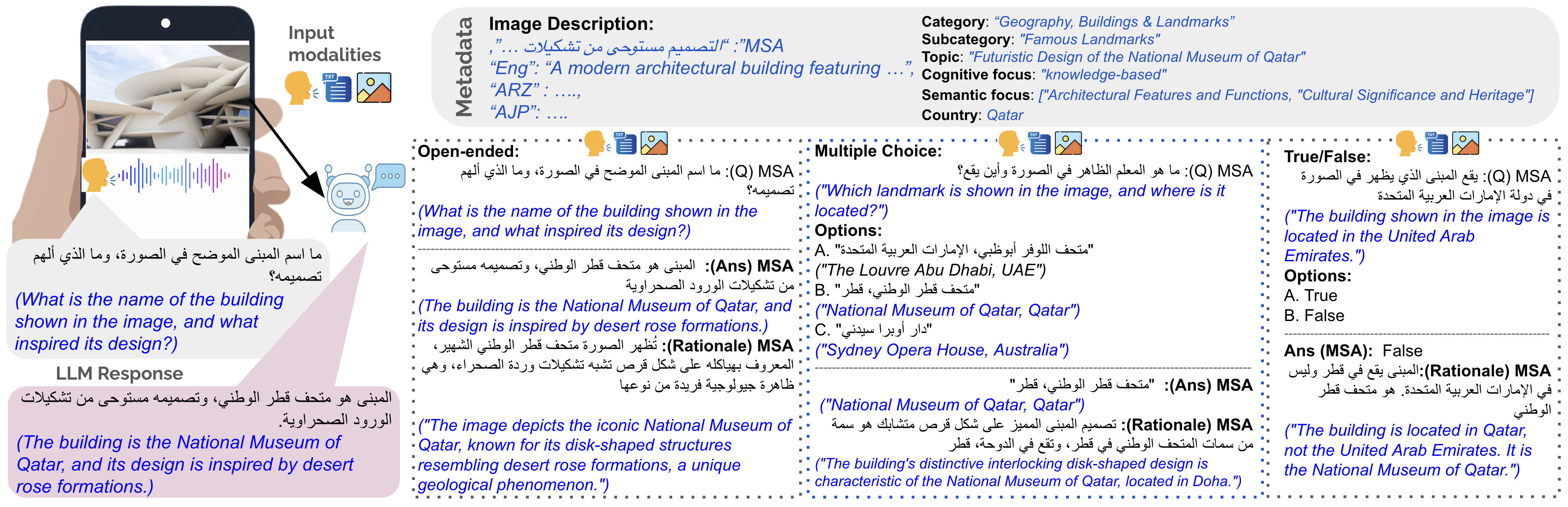}
\vspace{-0.3cm}
\caption{\oasis{} data sample, multimodal and multilingual QA around a culturally and visually grounded image. The correct answer requires visual evidence from the scene.}
\vspace{-0.5cm}
\label{fig:oasis_sample}
\end{figure*}

Humans experience the world through multiple senses: sight, hearing, touch, smell, and taste. This multi-sensory integration is fundamental to how humans understand the surroundings and communicate. As large language models (LLMs) evolve, it is increasingly important for them to process multiple modalities such as speech, text, and images to better mimic human interaction. For instance, when asking about an object, we often point to it while asking a question. In this setting, an AI assistant must reason over a multimodal triplet: \textit{visual information}, \textit{spoken information} (the question), and contextual knowledge needed for a \textit{culturally appropriate response} (Figure~\ref{fig:oasis_sample}).

Crucially, this contextual knowledge is not universal; it is shaped by culture and language. Successful multimodal reasoning therefore requires grounding perception and language in the cultural and linguistic context of the interaction, as gestures, phrasing, and interpretations vary across societies. However, current models are largely strong in high-resource contexts \citep{nayak-etal-2024-benchmarking,ananthram2025see}, leaving substantial room to better capture cultural and religious nuances in underrepresented languages. Existing resources remain limited: most focus on text-only or image--text settings, 
or introduce speech in isolation \citep{li2024culturepark,Rasheed2025PALO,Alam2024Maya,yue2024pangea}. As a result, they rarely capture the full \emph{object--question--culture} interaction and seldom include spoken queries grounded in visual context, both of which are essential for real-world multimodal reasoning.

This gap is particularly pronounced for languages such as Arabic. Cultural awareness is critical given its dialectal diversity \citep{ali2021connecting} and strong regional variation in traditions, religious expressions, and social norms. Linguistic variation spans MSA and numerous regional dialects such as Egyptian and Levantine, each reflecting distinct communicative practices. An AI assistant that ignores these differences, for example between everyday interactions in Morocco vs. Qatar, risks producing irrelevant or even inappropriate outputs. Advancing equitable AI therefore requires datasets that are multimodal, dialect-sensitive, and grounded in everyday cultural contexts.

\begin{wrapfigure}{r}{0.56\textwidth}
\vspace{-0.4cm}
\centering
\includegraphics[width=\linewidth]{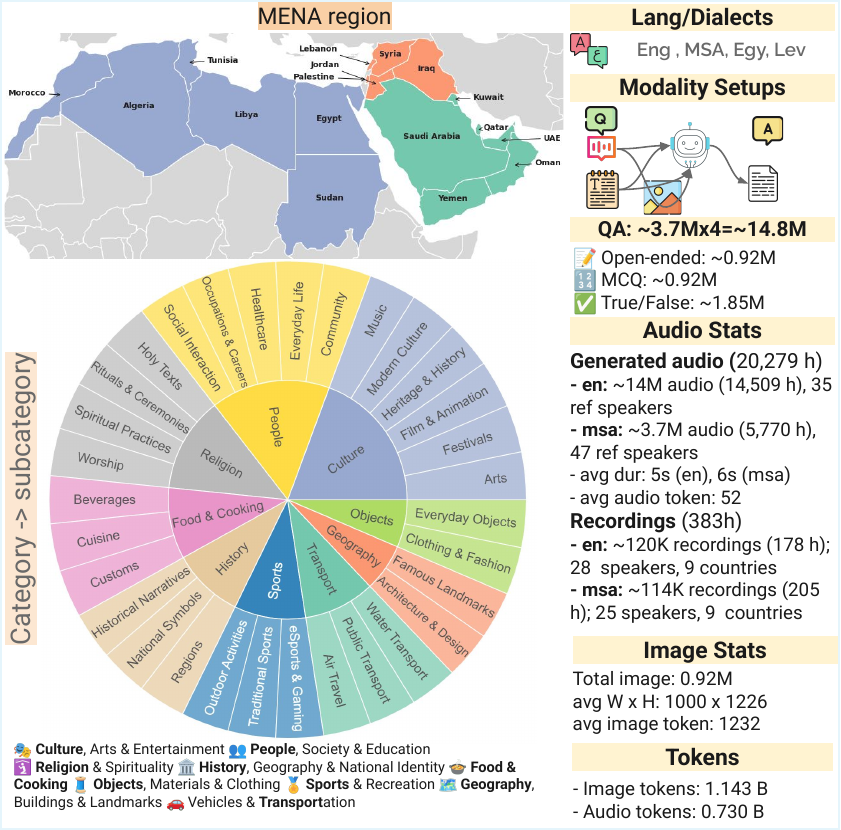}
\vspace{-0.4cm}
\caption{\oasis{} overview across 18 Arab countries, languages and dialects, modality setups, QA types, audio durations, token counts, and per-category distributions.}
\label{fig:dataset}
\vspace{-0.4cm}
\end{wrapfigure}

To address these limitations, we introduce \oasis{} (Figure \ref{fig:dataset}), a large-scale culturally grounded multimodal QA dataset for evaluating everyday visual, textual, and spoken reasoning. \oasis{} captures real-world interactions where correct answers depend on jointly reasoning over images, language, and cultural context. It spans multiple countries and dialects, supports diverse QA formats, and incorporates both text and speech inputs. To our knowledge, \oasis{} is the \textbf{\textit{first large multilingual triplet dataset}} of this scale (Table~\ref{tab:related-work}), providing a comprehensive benchmark for culturally grounded multimodal understanding.

\noindent\textbf{Task:}
We study culturally grounded spoken visual question answering -- given an everyday image and a user question provided either as text or speech, a model must produce or select a text answer grounded in both the visual scene and the relevant cultural context. We evaluate this task across text-only, speech-only, text+image, and speech+image setups, allowing us to distinguish reliance on language and world-knowledge priors from the use of instance-specific visual evidence.

Constructing a dataset of this kind is inherently challenging. It requires collecting images that are locally relevant to each region, generating natural queries that reflect how people actually ask questions, covering multiple dialects and languages, and ensuring alignment across modalities such as text and speech. Naively combining existing approaches such as translating English datasets, independently adding speech to vision corpora, or curating small multilingual resources fails to preserve this alignment. In practice, these pipelines often lose cultural specificity, break the connection between modalities, or do not scale.

\begin{figure*}[]
\centering
\vspace{-0.3cm}
\includegraphics[width=0.9\textwidth]{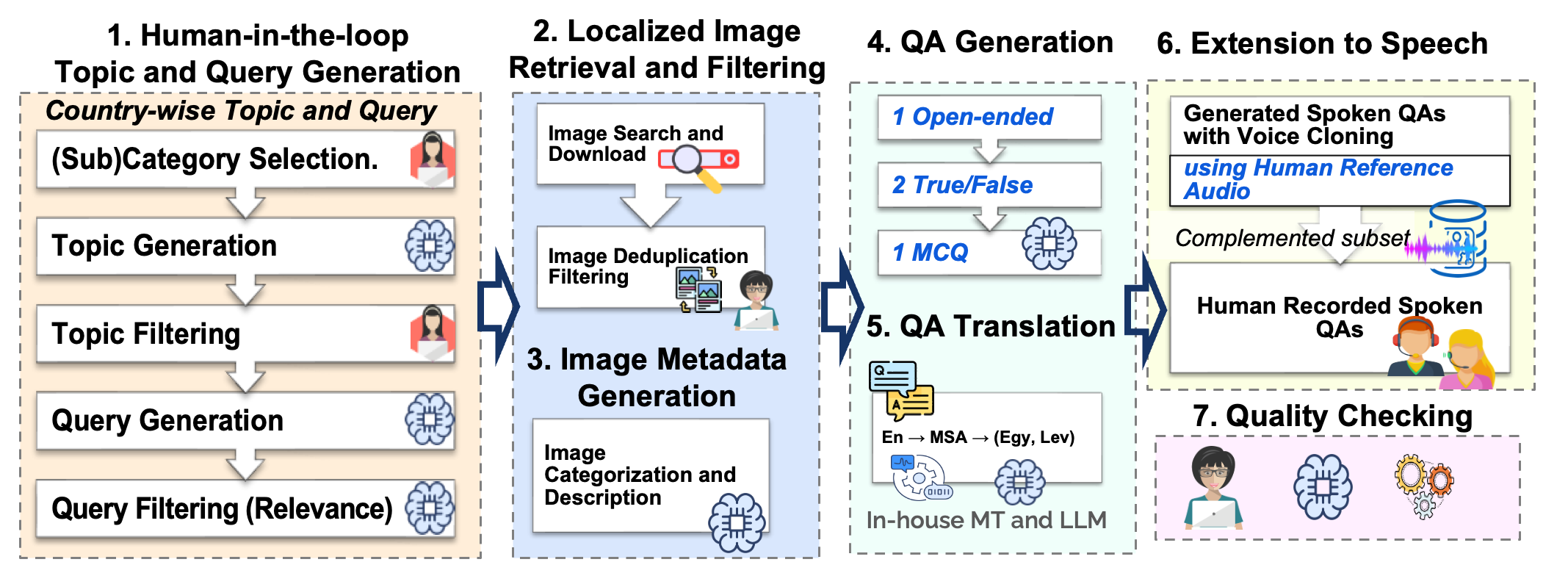}
\vspace{-0.2cm}
\caption{Pipeline for \emqa{} framework to create the \oasis{} dataset.}
\vspace{-0.3cm}
\label{fig:pipeline}
\vspace{-0.4cm}
\end{figure*}

These challenges motivate the need for a unified approach that can jointly handle localization, multimodal alignment, and quality control at scale. To this end, we develop \emqa{} (Figure \ref{fig:pipeline}), a language- and location-agnostic framework for constructing culturally grounded multimodal QA datasets. The framework integrates automated generation with human-in-the-loop validation, enabling scalable data creation while preserving cultural relevance. Although developed for \oasis{}, the framework is general and can be extended to new languages and regions.

Using \emqa{}, we build \oasis{} with images curated from 18 Arab countries. The dataset contains $\approx$\textbf{0.92M} real images and \textbf{14.8M} QA pairs in English, MSA, and Arabic dialects. Each image is paired with four QA instances: one open-ended question, one multiple-choice question, and two true or false questions. \oasis{} also includes $\approx$\textbf{20K hours of voice-cloned speech} and \textbf{383 hours of human-recorded speech}. These resources support four input settings: \textbf{text}, \textbf{speech}, \textbf{text+image}, and \textbf{speech+image}, enabling systematic evaluation of multimodal reasoning under realistic conditions.

\noindent\textbf{Contributions and findings:} We evaluate a suite of closed and open multimodal models on \oasis{} across all QA formats, and further fine-tune Qwen2.5-3B-Omni~\citep{Qwen2.5-Omni} to study whether \oasis{} can support adaptation in compact models. Our main contributions and findings are: 
\textit{(i)} \oasis{} provides a large-scale culturally grounded spoken visual QA benchmark that goes beyond image--text evaluation by jointly covering text, speech, images, languages, and dialects; 
\textit{(ii)} our evaluation shows that current models still rely on language and world-knowledge priors, but these shortcuts break down for examples requiring culturally situated visual evidence, while spoken inputs introduce additional brittleness from accents, acoustic variation, and transcription artifacts; and 
\textit{(iii)} \emqa{} enables scalable localized dataset construction, with \oasis{} supporting cross-modal adaptation in compact models and a preliminary transfer study suggesting applicability beyond Arabic.
Overall, \oasis{} and \emqa{} provide a foundation for studying culturally grounded multimodal understanding in real-world scenarios, and offer a scalable path toward building inclusive multimodal benchmarks beyond high-resource languages.

\section{\oasis{} Curation with \emqa{}}
\label{sec:data}

Existing resources lack the \textit{object–question–culture} triplet and \textit{spoken} queries, limiting real-world multimodal grounding for diverse language varieties like Arabic. To reduce the gap, we created \oasis{}, a multilingual, multimodal resource covering \textbf{18 Arabic-speaking countries} to ensure cultural and dialectal diversity. 
To create the dataset, we design a semi-automatic framework, \emqa{}, that ensures the content is grounded in local context and culture. The \emqa{} framework consists of seven stages (Figure~\ref{fig:pipeline}) and explicitly incorporates multiple human-in-the-loop steps to support cultural and contextual grounding.\footnote{We use LLMs at multiple stages, and provide all prompts used in the framework in Appendix~\ref{sec-app-prompts}.}

\subsection{Culturally Grounded \textbf{Topic \& Image Query Generation}}
We design a culturally grounded taxonomy, with \textit{9 categories} and \textit{31 subcategories}, inspired by \citep{schwenk2022okvqa,vayani2025all,nayak-etal-2024-benchmarking}. To collect country-specific real images, we use localized web search with user-oriented queries that reflect everyday information needs and its natural variation, including typos and grammatical errors. The process has two steps: \textit{topic generation and filtering}, followed by \textit{query generation and filtering}.

\textbf{Topic Generation and Filtering:} We use LLMs to generate 10 topics per subcategory, yielding 310 highly visual topics per country. The topics are then \textit{\textbf{manually reviewed and revised by knowledgeable contributors}} from each of the 18 countries to filter out generic or irrelevant cases and retain only those most suitable for image search. 

\textbf{Query Generation and Filtering:} Given the filtered topics, we generate large-scale naturalistic image-search queries using three LLMs:
GPT-4.1~\citep{openai2023gpt4}, Gemini-2.0-flash-001~\cite{team2023gemini},
and Claude-3.5-Sonnet.\footnote{\url{https://www.anthropic.com/news/claude-3-5-sonnet}} 
Using multiple models increases lexical and semantic diversity in the query pool. 
We merge and deduplicate the generated queries, obtaining approximately 6,100 unique candidate queries per country.
As image search over thousands of queries is expensive, and because not all generated queries are equally culturally or geographically relevant, we apply a relevancy filtering step. We prompt \texttt{GPT-4o} to assign each query a cultural-relevancy score \(R_{\text{LLM}} \in [1,100]\), reflecting its location and cultural fit given the country, category, subcategory, and topic context. We retain only queries with \(R_{\text{LLM}} \ge 80\). This threshold was chosen based on manual inspection to balance coverage and specificity.
After filtering, the average number of retained queries is 176.1~$\pm$~13.5 across countries 
and 174.3~$\pm$~14.4 across categories 
(see statistics in Appendix Table~\ref{tab:ablation_topic_query}).
The variation across subcategories is expected: highly visual and country-specific subcategories, such as \textit{Famous Landmarks} and \textit{Heritage \& History}, retain more queries, while less visually grounded subcategories, such as \textit{eSports \& Gaming}, retain fewer. This variation reflects differences in the availability of digital visual content across countries.

\vspace{-0.2cm}
\subsection{Country-Localized \textbf{Image Retrieval} and \textbf{Filtering} }
\label{sec: image_Retrieval}
\textbf{Image Retrival:} Using the filtered relevant queries for each country, we retrieve images through country-localized Google Custom Search. The search uses locale settings, safe search, and license filters restricted to \texttt{cc\_publicdomain}, \texttt{cc\_attribute}, and \texttt{cc\_sharealike}. For each query, we collect the top 20--40 results and keep only images that meet minimum-resolution and standard file-type requirements. We also store source metadata for each image, including the URL, query, country, category, subcategory, and topic. This step retrieves $\sim$\textbf{4.3M} images from \textbf{18 Arab countries}.

\textbf{Image Filtering:}
We first apply URL-based deduplication, which reduces the pool to about \textbf{2.4M} unique entries. Due to timeouts and broken links, about \textbf{1.4M} images are successfully downloaded. We then remove exact and near duplicates using features extracted from a fine-tuned ResNet18~\citep{he2016deep}. Images with highly similar embeddings are treated as duplicates and removed. This produces a final deduplicated set of $\sim$\textbf{1.35M} unique images.

\subsection{Metadata: Image Categorization and Description Generation}
After deduplication, we manually inspected the images through a web portal we developed. The collection still included non-representative items, such as advertisements, charts, screenshots, infographics, illustrations, and images with text overlays. We filtered these cases by using an LLM to assign each image one of the following category labels: \textit{photograph}, \textit{advertisement}, \textit{illustration}, \textit{chart/infographic}, \textit{meme/text overlay}, \textit{screenshot/UI capture}, or \textit{other}. We also used an LLM to generate a image description conditioned on the image, category, and subcategory, supporting later QA generation. Finally, we retained only images labeled as \textit{photograph}, yielding  $\approx$\textbf{1.30M} images.
\vspace{-0.3cm}
\subsection{QA Generation}
Developing multimodal and multilingual QA resources is costly and time-consuming specially when manually created \citep{changpinyo-etal-2023-maxm}. Recent work shows that LLMs and VLMs can support scalable QA generation~\citep{zhang2025automated,vayani2025all,yue2024pangea,urailertprasert-etal-2024-sea}. We follow this direction, while keeping \textit{\textbf{human evaluation central to quality validation}}.
For each image, we generate four questions across three formats: one open-ended (OE) question, one multiple-choice (MCQ) question, and two true/false (T/F) questions. \textbf{Open-ended} questions test free-form grounded reasoning, \textbf{MCQs} enable objective evaluation with plausible distractors, and \textbf{T/F} questions help measure factuality and hallucination~\citep{li2023evaluating}. 
We include two T/F questions per image to measure both false positives (asserting details not present) and false negatives (missing present content).
We also assign each question a semantic and cognitive label. The semantic labels cover 11 categories, such as location, architecture, and cultural heritage, and are derived by clustering annotator-written questions. The cognitive labels distinguish \textit{knowledge-based} from \textit{commonsense-based} questions, following prior VQA benchmarks~\citep{Marino2019OKVQA,schwenk2022okvqa,Zellers2019VCR}. Details are provided in Appendix~\ref{sec-app:semanticNcogni}.
Each QA generation is conditioned on the image, its description, and its category and subcategory. For each generated question, we store the answer, rationale, semantic label, and cognitive label. Images that trigger content filters are removed. Overall, this process yields \textbf{3.7M English QA pairs} from \textbf{0.92M images}.

\vspace{-.3cm}

\subsection{Translation}

To create multilingual and dialectal QA pairs, we used machine translation. For MSA, we translated the English QA pairs with an in-house LLM-based system, which achieved an average BLEU score of 25.11 across 11 datasets, outperforming Google Translate at 19.62. For dialectal Arabic, we compared GPT-4.1 with a two-step in-house English$\rightarrow$MSA$\rightarrow$dialect pipeline. Since both BLEU scores and native-speaker evaluations favored GPT-4.1 (Table~\ref{tab:dialectal-bleu-scores}, Appendix~\ref{apd:mt-results}), we used GPT-4.1 for dialectal translation.

\subsection{Building Spoken QA Resources}

The Spoken-QA task aims to support natural LLM interaction in a speech-in, text-out setting, where models process spoken questions and use acoustic cues. Since large-scale natural spoken QA data is difficult to collect, we use XTTS-v2~\citep{casanova2024xtts} to synthesize English and MSA speech through voice cloning with real-speaker reference audio. We also collect a \textit{high-quality human-recorded benchmark set}. Due to the limited availability of robust Arabic dialectal TTS models, we focus on English and MSA.

\noindent
\textbf{Generated Spoken QA:}
We first constructed reference voice resources for speech generation. For English, we sampled 5--8 second segments from LibriTTS~\citep{zen2019libritts}, selecting 10 utterances per speaker from 35 speakers and obtaining 337 segments. For Arabic, we built a reference voice bank from QASR~\citep{mubarak2021qasr} and ADI17~\citep{shon2020adi17}, covering MSA and several dialects across 47 speakers. Applying the same 5--8 second duration criterion, we obtained 389 segments from 40 QASR speakers and 69 manually reviewed segments from 7 ADI17 speakers. For each question, we synthesized three English audio samples and one MSA audio sample using randomly selected reference speakers.

\noindent
\textbf{Human-Recorded Spoken QA:}
To complement synthetic speech, we collect a human-recorded spoken QA benchmark in English and MSA with country-specific questions (a subset of test set). Native Arabic speakers and fluent English speakers record all question types using our in-house platform under natural conditions. This produces about \textbf{383 hours} of speech, covering 234K spoken questions from 42 distinct speakers. 
This benchmark provides a realistic reference for evaluating spoken QA.

\vspace{-0.3cm}
\subsection{Quality Assessment}
\noindent
\textbf{Manual annotation:} We manually annotated a large portion of the test-set QAs across all countries. QAs were rated for clarity on a five-point Likert scale, except T/F answers, which used a three-point scale. Annotators provided justifications for low scores. Rationales were also evaluated along two combined dimensions: (\textit{i}) clarity \& informativeness, and (\textit{ii}) plausibility \& faithfulness \citep{wang_evaluating_2023,huang_chain_2023,agarwal2024faithfulness}.
Annotation was conducted by native Arabic speakers using a dedicated web interface, guidelines, and expert supervision, with quality checks. The detailed annotation guidelines are provided in Appendix~\ref{app_section_annotation_guidelines}. In total, 26,473 samples from 14 countries were annotated, yielding $\sim$212K annotations ($26,473 \times 4$ QAs $\times 2$ annotators).

\textbf{Annotation agreement:} All QA evaluation metrics were rated on Likert scales, with the average of two annotators reported per item. To measure \textit{inter-annotator} agreement on ordinal scales, we used the $r^*_{wg(j)}$ index \citep{james1984estimating}.
In Table~\ref{tab:qa_annotation_quality} (Appendix \ref{sec-app:agreement}), we report both the human and LLM-based annotation agreement across open-ended, MCQ, and T/F types (LLMs evaluated on 18 and human on 14 countries). Our results show near-perfect agreement in answer consistency for T/F questions (LLMs: 0.97, humans: 0.93). MCQ scores are also very strong (LLMs: 0.87, humans: 0.96).
The $r^*_{wg(j)}$ scores for answer quality in open-ended questions are 0.79 and 0.95 for LLMs and humans, respectively.

\textbf{LLM-based annotation:} We further employed Gemini-2.5-Pro~\citep{team2023gemini} and Llama-4-Scout-17B-16E-Instruct\footnote{\url{https://huggingface.co/meta-llama/Llama-4-Scout-17B-16E-Instruct}} to complement the human annotation tasks~\citep{zheng2023judging}. Using the same guidelines and inputs (image and description), the models generated annotations for the complete test split.

\noindent \textbf{LLM-Human Agreement:}
We assessed alignment between LLM annotations and human judgments across all QA formats. The correlations for question quality, answer quality, rationale plausibility \& faithfulness, and rationale clarity \& informativeness are 0.93, 0.87, 0.86, and 0.93, respectively, showing that LLMs can serve as reliable complementary annotators for quality assessment.

\noindent \textbf{Generated Spoken QA Quality:} We evaluate generated English and MSA audio using three metrics: \textit{(i)} Word Error Rate (WER) for transcription accuracy, using Whisper-small~\citep{radford2023robust} for English and Fanar\footnote{A publicly accessible ASR API: \url{https://fanar.qa/}.} for MSA; \textit{(ii)} Speaker Cosine Similarity (SpkCos) for speaker consistency, using embeddings from \texttt{spkrec-ecapa-voxceleb}~\citep{ravanelli2021speechbrain}; and \textit{(iii)} NISQA~\citep{mittag2021nisqa} for perceptual speech quality, including naturalness and distortion. For human recordings, we use WER with the same ASR systems. As shown in Table~\ref{tab:audio_quality_comparison} in Appendix, generated English speech achieves high perceptual quality (NISQA: 4.33), while generated MSA shows moderate quality (NISQA: 3.68). Similar SpkCos scores across languages indicate consistent speaker similarity.

\noindent \textbf{Human-Recorded Spoken QA Quality:}
For human-recorded speech, both English and Arabic show higher WER than synthetic speech (Table~\ref{tab:audio_quality_comparison}). This reflects the natural variability of real-world speech, including accents, background noise, and recording conditions, making this split a more realistic and challenging robustness benchmark. Arabic recordings show higher WER, mainly due to phonological and dialectal variation, noisy acoustic conditions, and cross-script transliteration.

\vspace{-0.3cm}
\subsection{Data Analysis and Stats}

Figure~\ref{fig:dataset} summarizes the key statistics of \oasis{}, which spans 18 MENA countries across text, speech, and image modalities. The dataset includes 3.7M QAs (open-ended, MCQ, T/F). In four language varieties this results in $\sim$14.8M QA pairs. It exhibits balanced country coverage, with totals ranging from $\sim$36K (Qatar) to $\sim$64K (Morocco) (median $\sim$53K). Category representation is generally consistent (1–2.3K images per country), with notable cultural peaks (e.g., \textit{Traditional \& Regional Cuisines}) and lows (e.g., \textit{Clothing \& Fashion}).
Detailed country and subcategory statistics are provided in Appendix Tables~\ref{tab:country_cat_tableA} and \ref{tab:country_cat_tableB}.

\textbf{Audio:} The \textit{synthesized audio} covers $\sim$20,279 hours in English and MSA, with average QA durations of 5s and 6s, respectively, closely matching human recordings. The \textit{human recordings} total $\sim$383 hours from 42 speakers across 12 countries. Additional audio statistics are shown in Figure~\ref{fig:dataset}. 

\textbf{Images:} The image set %
has an average resolution of $1000\times1226$ px (width 372–2185 px, height 453–2415 px), confirming high visual quality. Tokens are computed tile-wise: each $512\times512$ tile contributes 85 base tokens plus 170 per tile, from which we derive both per-image and global totals.

\textbf{Data split:} Table~\ref{tab:data_split_by_country} (Appendix) reports the train, dev, and test splits for each country. Splits were created via subcategory-wise stratification, with $\sim$3.76\% allocated to dev and test each (about 2K samples per country, $\sim$35K total per split), and the remainder allocated to training.

\section{Experimental Setup} \label{sec: experiments}

\label{sec:experiments}

\paragraph{Task definition:}
We evaluate models on culturally grounded spoken visual question answering. Each instance consists of an image $I$, a question $q$, and an answer $a$. The question is provided either as text $q_{\text{text}}$ or speech $q_{\text{speech}}$, and the model always produces a text output. Depending on the QA format, the output is evaluated as a free-form answer, a selected multiple-choice option, or a true/false decision. We consider four core input regimes: text-only ($T$: $q_{\text{text}}$), speech-only ($S$: $q_{\text{speech}}$), text+image ($T{+}I$: $(I, q_{\text{text}})$), and speech+image ($S{+}I$: $(I, q_{\text{speech}})$). The text-only and speech-only regimes measure how much models can answer from linguistic and world-knowledge priors alone, while the image-grounded regimes test whether models can use instance-specific visual evidence together with culturally situated context. We additionally consider ASR transcript inputs ($T_r$) as a diagnostic alternative to raw speech.

\textbf{Benchmarking Models and Setups:} We evaluated six models from closed and open families: \textbf{GPT-4.1}~\citep{openai2023gpt4}, \textbf{GPT-5}~\citep{singh2026openaigpt5card}, and \textbf{Gemini-2.5-Pro} (closed), and \textbf{Qwen-2.5-7B}, \textbf{Qwen-2.5-3B}~\citep{wang2024qwen2}, and \textbf{Phi-4}~\citep{microsoft2025phi4minitechnicalreportcompact} (open). This selection covers capabilities from frontier models to smaller, more accessible open-source ones.
All models are evaluated in a zero-shot setting under four input configurations: text (\textit{T}), speech (\textit{S}), text+image (\textit{T+I}), and speech+image (\textit{S+I}), with outputs are always generated as text. We evaluate three task formats per item: open-ended (OE) generation, MCQ, and two true/false variants (TF1 and TF2). Since the \textbf{\textit{human-recorded}} split covers only part of the test set, about 383 hours of speech, \textit{S} and \textit{S+I} evaluations are limited to items with open-ended and true/false recordings (Appendix Table~\ref{tab:recordings_evaluation}).
We ran experiments based on each model's supported inputs, resulting in \textbf{108 distinct configurations} across models, modalities, and languages verities (See Table~\ref{tab:exp_modal_combo} in Appendix).

\textbf{Model Fine-Tuning:}
We fine-tuned Qwen2.5-Omni-3B~\citep{Qwen2.5-Omni} on different input configurations: \textit{S+I}, \textit{T+I}, \textit{T}, and \textit{S} only. We also explored using ASR transcripts (\textit{T$_r$}) instead of raw speech. To ensure fair comparison, we used the same prompt template and response schema for each task, and kept decoding settings fixed across models, with temperature set to 0, top-\textit{p} set to 1.0, and a fixed maximum output length.

The full \oasis{} training split contains $\approx$859.6K images, with four questions per image. We construct training instances across four language varieties: English, MSA, Egy, and Lev Arabic. English and MSA include four input configurations each, while Egy and Lev Arabic include two configurations each. This results in  $\approx$41.26M possible training datapoints.
Due to compute constraints, we fine-tune on 6.67\% of the full training set, corresponding to  $\approx$2.75M datapoints. We use LoRA\citep{hu2022lora} with rank 16 and $\alpha=32$, a learning rate of $2\times10^{-5}$, a maximum sequence length of 3072 tokens, and train for one epoch.

\textbf{Evaluation and metrics:}
We evaluate models on the \oasis{} test set using standard QA metrics. For open-ended questions, we report BERTScore~F1~\citep{zhangbertscore} and GPT-4.1 judge scores, following MT-Bench~\citep{zheng2023judging}, with a 1--10 rubric for helpfulness, relevance, accuracy, and faithfulness. We also use human annotations on a subset to assess agreement with the LLM judge. For MCQ and T/F questions, we report accuracy. Finally, we test the generalization of the fine-tuned model on Arabic subset of ALM-Bench~\citep{vayani2025all}.

\section{Results and Discussion}

We organize the results around three questions central to culturally grounded multimodal QA: 
\textit{(i)} how much visual evidence improves answer quality, 
\textit{(ii)} whether visual grounding reduces language and dialect-specific difficulty, and 
\textit{(iii)} how robust models are when the query is delivered through speech rather than clean text. 
We begin with open-ended QA as it best reflects realistic user interaction where models must generate faithful, contextually appropriate answers rather than choose from fixed options. 
Table~\ref{tab:results_text_image_oe} reports open-ended performance across modalities, languages, and dialects, while MCQ and T/F results are provided in Table~\ref{tab:results_text_image} in the Appendix as complementary checks of visual grounding and factual consistency.

\textbf{Visual grounding shifts the bottleneck:} Providing the image as context yields large, consistent gains across all models. For open-ended answers, LLM-as-judge scores improve significantly (about $1$--$2{+}$ points), while BERTScore gains are modest. This indicates that visual evidence resolves recognition/grounding, shifting the primary bottleneck to faithful answer generation. 

\begin{wraptable}{r}{0.6\textwidth}
\vspace{-0.2cm}
\centering
\caption{Evaluation results across languages and modalities. \textbf{F1} = BERTScore F1, \textbf{Judge} = LLM-as-judge score, \textbf{T} = text, \textbf{T+I} = text+image, and Gemini = Gemini-2.5-pro. Underlining marks the best \textbf{T}-only result per language, and bold marks the best \textbf{T+I} result for OE Judge, MCQ, and T/F.}
\label{tab:results_text_image_oe}
\vspace{-0.2cm}
\setlength{\tabcolsep}{2pt}
\renewcommand{\arraystretch}{0.92}
\scalebox{0.70}{
\begin{tabular}{llcc|cc|cc|cc}
\toprule
\multirow{2}{*}{\textbf{Model}} & \multirow{2}{*}{\textbf{Mod.}}
& \multicolumn{2}{c|}{\textbf{English}}
& \multicolumn{2}{c|}{\textbf{MSA}}
& \multicolumn{2}{c|}{\textbf{Egyptian}}
& \multicolumn{2}{c}{\textbf{Levantine}} \\
\cmidrule(lr){3-4} \cmidrule(lr){5-6} \cmidrule(lr){7-8} \cmidrule(lr){9-10}
& & \textbf{F1} & \textbf{Judge}
  & \textbf{F1} & \textbf{Judge}
  & \textbf{F1} & \textbf{Judge}
  & \textbf{F1} & \textbf{Judge} \\
\midrule
GPT-4.1   & T   & 0.60 & 6.26 & 0.58 & 6.36 & 0.56 & 6.07 & 0.57 & 6.41 \\
        & T+I & 0.73 & 8.60 & 0.62 & 8.36 & 0.62 & 8.30 & 0.62 & 8.39 \\
GPT-5   & T   & 0.61 & \underline{6.39} & 0.55 & \underline{6.39} & 0.53 & \underline{6.18} & 0.55 & 6.31 \\
        & T+I & 0.66 & \textbf{8.46} & 0.57 & \textbf{8.10} & 0.55 & \textbf{7.86} & 0.57 & \textbf{8.03} \\
Gemini  & T   & 0.57 & 5.50 & 0.54 & 5.69 & 0.51 & 5.42 & 0.52 & 5.60 \\
        & T+I & 0.63 & 7.14 & 0.56 & 6.90 & 0.52 & 6.62 & 0.55 & 6.85 \\ \midrule
Qwen-7B & T   & 0.57 & 5.11 & 0.53 & 4.45 & 0.48 & 4.07 & 0.49 & 4.23 \\
        & T+I & 0.64 & 5.10 & 0.55 & 4.45 & 0.49 & 5.70 & 0.51 & 5.76 \\
Phi-4   & T   & 0.55 & 5.01 & 0.51 & 3.71 & 0.46 & 2.94 & 0.47 & 3.24 \\
        & T+I & 0.59 & 6.22 & 0.51 & 4.16 & 0.46 & 3.56 & 0.48 & 3.76 \\
Qwen-3B & T   & 0.54 & 4.78 & 0.52 & 3.84 & 0.47 & 3.21 & 0.49 & 3.60 \\
        & T+I & 0.50 & 5.27 & 0.52 & 4.91 & 0.45 & 4.18 & 0.46 & 4.32 \\
\midrule
FT Qwen-3B
        & T   & \underline{0.73} & \underline{6.39} & \underline{0.64} & 5.85 & \underline{0.64} & 5.85 & \underline{0.64} & 5.96 \\
        & T+I & \textbf{0.78} & 8.29 & \textbf{0.67} & 7.47 & \textbf{0.67} & 7.35 & \textbf{0.67} & 7.56 \\
\bottomrule
\end{tabular}}
\vspace{-0.25cm}
\end{wraptable}

\textbf{\textit{Why visual grounding help?}}
In text-only settings, MSA is challenging because of rich morphology, orthographic variation, and data sparsity, which can make referents and attributes harder to resolve. Adding images improves performance across English and Arabic variants, with especially strong gains for Arabic in classification-style tasks such as MCQ and TF2. For open-ended QA, moving from T to T+I gives only a modest gain in BERTScore F1 across models and language variants (+0.03~$\pm$~0.03), but a much larger gain in LLM-as-judge scores (+1.32~$\pm$~0.65), as shown in Table~\ref{tab:oe_gain_all_models}. This suggests that images improve semantic grounding and answer quality more than surface-form overlap. The remaining English--MSA gap is more likely due to generation-level issues, such as word choice and agreement.

\textbf{\textit{Does grounding reduce language and dialect difficulty?}}
Dialectal inputs are more challenging in text-only settings because they are low-resourced and show greater linguistic variation than MSA. Visual grounding helps reduce this burden by anchoring the context and constraining the set of plausible answers. This leads to higher accuracy on constrained tasks and substantially narrows the gap between dialects and MSA. Levantine nearly converges with MSA, while the small remaining MCQ gap for Egyptian likely reflects dialect-specific generation or option-selection issues.

\textbf{\textit{Beyond language priors.}}
Naturalistic multimodal QA may exhibit non-trivial text-only performance due to language and world-knowledge priors \cite{goyal2017making,yue2024mmmu,schwenk2022okvqa}. Rather than treating this as a confound to eliminate, we measure the marginal value of instance-specific visual evidence. We identify 7,809 strong image-dependent samples through a separate annotation round with three annotators. On this subset, adding images improves judge accuracy by +3.10 overall, indicating that visual grounding contributes substantial information beyond priors, particularly for culturally grounded and context-dependent reasoning.

\textbf{Speech {\em vs.} Text Modalities:}
We next examine whether the input channel affects performance for the same underlying query. Clean text (\textbf{T}) consistently outperforms both ASR transcripts (\textbf{Tr}) and raw speech (\textbf{S}) (Figure~\ref{fig:judge_merged_msa}), reflecting noise in the speech input pipeline. Raw speech adds acoustic variation, background noise, and speaker differences, while transcripts reduce acoustic uncertainty but introduce ASR errors and normalization artifacts. Thus, \textbf{Tr} remains less reliable than clean \textbf{T}. Detailed results are provided in Appendix~\ref{sec-app: details_rslt}, Table~\ref{tab:speech_combination_modal_results}.

\textbf{\textit{Does visual grounding reduce the speech-channel penalty?}}
Adding the image substantially reduces the gap between speech-based inputs and clean text. The visual signal provides channel-independent evidence about objects, entities, attributes, and scene context, helping the model recover from acoustic and ASR noise. The largest gains are observed for \textbf{S}$\rightarrow$\textbf{S+I}, where the image restores missing cues and constrains the answer space. This suggests that visual grounding acts as a modality equalizer, especially when the question is provided through a noisy speech channel.

\begin{wrapfigure}{r}{0.48\textwidth}
\vspace{-0.3cm}
\centering
\captionsetup{width=\linewidth}
\begin{tikzpicture}
\begin{axis}[
width=\linewidth,
height=4.5cm,
ymin=1, ymax=8.0,
xtick={1,2,3,4,5,6},
xticklabels={T,T+I,Tr,Tr+I,S,S+I},
x tick label style={font=\scriptsize},
y tick label style={font=\scriptsize},
ylabel={Judge},
ylabel style={font=\scriptsize},
ylabel shift={-5pt},
legend style={
    at={(0.5,1.02)},
    anchor=south,
    legend columns=2,
    font=\scriptsize,
    draw=none
},
grid=major,
grid style={dotted, gray!35},
mark size=1.5pt,
line width=0.8pt
]

\addplot+[mark=o, color=gray!80!black] coordinates
{(1,3.94) (2,4.90) (3,3.16) (4,3.73) (5,2.15) (6,2.95)};

\addplot+[mark=square*, color=orange!85!black] coordinates
{(1,4.48) (2,6.28) (3,3.64) (4,5.38) (5,2.79) (6,4.50)};

\addplot+[mark=triangle*, color=blue!75!black] coordinates
{(1,5.71) (2,6.69) (3,4.83) (4,6.18) (5,2.87) (6,5.92)};

\addplot+[mark=diamond*, color=teal!70!black] coordinates
{(1,5.93) (2,7.60) (3,5.19) (4,7.23) (5,4.97) (6,7.22)};

\legend{Qwen-3B, Qwen-7B, Gemini, FT-Qwen-3B}
\end{axis}
\end{tikzpicture}
\vspace{-0.25cm}
\caption{MSA judge scores across input modalities.}
\label{fig:judge_merged_msa}
\vspace{-0.25cm}
\end{wrapfigure}

\textbf{Closed {\em vs.} Open models:} Closed models perform 
better than open models even in text-only settings, often achieving scores closer to the gold standard (e.g., GPT-4.1 {\em vs.} Qwen-2.5-7B on English and MSA MCQ). This suggests closed models leverage world knowledge and priors or benefit from broader pretraining and instruction tuning to make educated guesses and narrow the hypothesis space without visual evidence.

\textbf{Effect of Fine-tuning:}
We fine-tune Qwen-3B on the multimodal training data to test whether a smaller open model can better align text, speech, and image inputs. As shown in Figure~\ref{fig:judge_merged_msa}, fine-tuning improves performance across input channels, especially for raw speech (\textbf{S}) and ASR transcripts (\textbf{Tr}), with larger gains when images are available. In English (Appendix Figure \ref{fig:judge_merged}), \textbf{S+I} and \textbf{Tr+I} approach clean \textbf{T+I} performance, showing that fine-tuning helps the model use visual evidence under noisy inputs. It also reduces the \textbf{Tr+I} instability seen in the base model, suggesting better vision--text fusion and robustness to ASR artifacts. Overall, fine-tuning improves cross-modal alignment and makes the small model more competitive for audio-based QA.

\textbf{\textit{Robustness to audio noise.}}
We also evaluate robustness to noisy audio by adding realistic background noise at 0--10 dB SNR to the human-recorded Algerian MSA test set.
For open-ended QA, Qwen-3B-FT shows only a small drop, from 7.29 to 7.04, with similar trends for Qwen-3B and T/F QA. Regenerating the same samples with the original speakers as references (for voice cloning) slightly improves performance over human recordings (7.47 vs. 7.29), suggesting that the generation pipeline preserves task-relevant variation.

\textbf{\textit{LLM-Judge Agreement with Humans:}}
To validate the reliability of our open-ended evaluation, we compare GPT-4.1 judge scores with human judgments on 800 QA items, each annotated by three annotators. The agreement is high ($r_{wg}=0.94$), indicating strong alignment between the LLM judge and human evaluators. This supports the use of GPT-4.1 for scalable open-ended QA evaluation, consistent with prior work on LLM-based judging~\cite{zheng2023judging}.

\textbf{\textit{Generalization beyond \oasis{}:}}
To test whether the fine-tuned model generalizes beyond our benchmark, we evaluate it on the Arabic subset of \textbf{ALM-Bench}~\cite{vayani2025all} and its translated variant, covering 770 images. We compare the Qwen-3B base model and our Omni-3B-FT model.
The fine-tuned model improves over the reported Qwen-2-VL baseline by +0.11 absolute accuracy on average, with gains of +0.10, +0.10, and +0.13 for Saudi, Egyptian, and Emirati Arabic, respectively (Appendix Table~\ref{tab:results_alm_bench_acc_llm_judge}). Additional results by question type are provided in Appendix Table~\ref{tab:results_alm_bench}.

\textbf{\emqa{} extends beyond Arabic:} To assess whether \emqa{} generalizes beyond OASIS, we conduct a preliminary transfer study for Bangla in Bangladesh, a low-resource setting outside the main scope of this work. Using the same core pipeline, the Bangla instantiation yields $\sim$1.5K localized search queries, $\sim$70K candidate images, $\sim$6K retained images after filtering, and $\sim$64K QA pairs. This provides initial evidence that \emqa{} can be adapted to new linguistic and cultural contexts. We treat this as a feasibility check rather than a full benchmark: Bangla audio generation, human recording, quality validation, and model evaluation are left for future work.

\textbf{Keys Findings:}
Our results show that image-centric questions are best answered when models ``\textit{see what the user sees}.'' Visual grounding provides large and consistent gains across models, languages, and dialects, while also narrowing cross-lingual and dialectal gaps. These gains are %
important for culturally grounded questions, where the image provides context that is difficult to recover from text alone. Remaining errors suggest the bottleneck shifts from visual recognition to faithful, well-calibrated answer generation.
These findings highlight \oasis{} as a large-scale benchmark for culturally grounded multimodal QA across text, image, and speech, and \emqa{} as a scalable framework for building localized QA resources beyond a single language or region. Together, they provide a foundation for studying how multimodal models handle everyday, culturally specific, and linguistically diverse user interactions.

\vspace{-0.3cm}

\section{Related Work}
\label{sec:related work}

\textbf{Omni -- Large Multimodal Models:}
Recent ``omni'' LMMs unify text, vision, audio, and video within a single architecture. Examples include \textsc{Qwen2.5-Omni}~\citep{Qwen2.5-Omni}, Phi-4~\citep{abdin2024phi}, and closed models such as Gemini~\citep{team2023gemini}. These systems handle diverse inputs and generate text or speech. For instance, \textsc{Qwen2.5-Omni} employs a \emph{Thinker–Talker} design with time-aligned multimodal encoding, achieving strong results on OmniBench~\citep{li2024omnibench}. Microsoft’s \textsc{Phi-4-Multimodal} extends the Phi-4 recipe to vision–audio–text with multilingual reasoning, while earlier efforts such as \textsc{Kosmos-2}~\citep{Peng2023Kosmos2} foreshadowed this omni-modal direction. These models natively support our input regimes ($T$, $T{+}I$, $S$, $S{+}I$) with text outputs, but multilingual coverage, especially for Arabic, remains limited.

\textbf{Frameworks and datasets:}  
Recent work has addressed culturally grounded multimodal resources, primarily through: \textit{(i)} translating English corpora (PALO ~\citep{Rasheed2025PALO}), \textit{(ii)} curating multilingual resources (Maya~\citep{Alam2024Maya}, Pangea~\citep{yue2024pangea}), or \textit{(iii)} adding speech to vision datasets (SPEECH-COCO). 
In Table \ref{tab:related-work} (Appendix), we summarize the most notable existing QA datasets and compare them with ours.

\section{Conclusions and Future Work}
\label{sec:conclusions}

This paper introduces \oasis{}, a large-scale culturally grounded multimodal QA dataset for evaluating everyday visual, textual, and spoken reasoning. \oasis{} spans 18 Arabic-speaking countries and includes approximately 0.92M real images, \textbf{14.8M QA pairs}, and \textbf{3.7M spoken questions}, along with approximately \textbf{20K hours of voice-cloned speech} and \textbf{383 hours} of \textbf{human-recorded} speech from \textbf{42 speakers}. To our knowledge, \oasis{} is the largest tri-modal resource focused on English and Arabic varieties. \oasis{} is built using \emqa{}, a language- and location-independent framework for scalable multimodal data creation with human-in-the-loop validation. Our experiments show that visual grounding consistently improves performance across models, languages, and dialects, while fine-tuning on \oasis{} strengthens cross-modal alignment, especially in text+image, transcript+image, and speech+image settings. These results establish \oasis{} as a comprehensive benchmark and training resource for culturally grounded multimodal QA. Future work will scale training to the full dataset and extend the framework to additional languages and regions.

\begin{ack}
\end{ack}

\bibliography{bibliography/bibliography}
\bibliographystyle{plainnat}

\appendix

\section*{Appendix}

\startcontents[appendix]

\section*{Appendix Table of Contents}
\printcontents[appendix]{}{1}{\setcounter{tocdepth}{2}}

\section{Details of the \emqa{} and \oasis{}}

\subsection{Topic \& Image Query Generation}
\label{sec-ablation-topic-query}

\begin{figure*}
    \centering
    \includegraphics[width=0.75\linewidth]{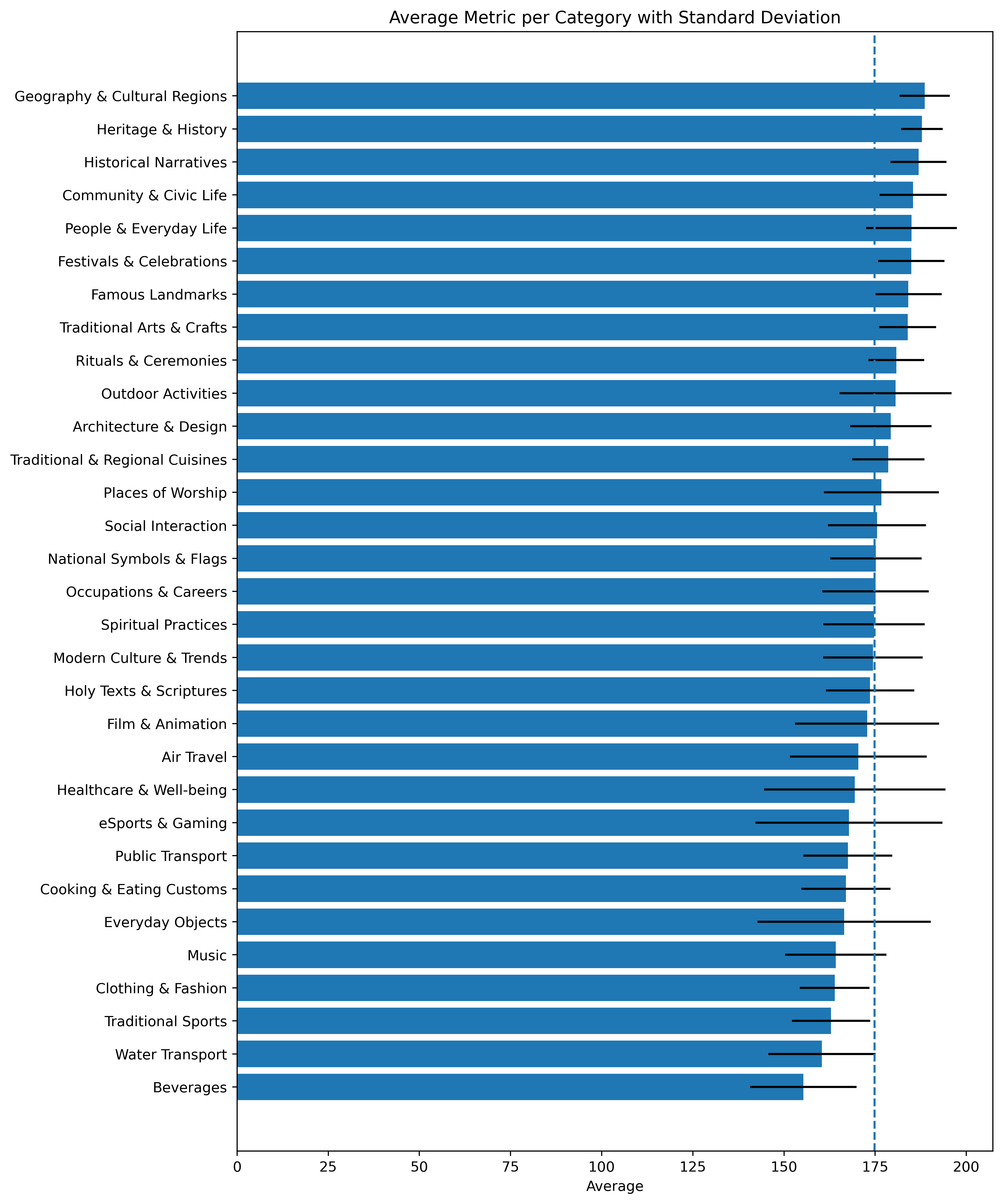}
    \caption{Average per category with standard deviation as error bars. Categories are sorted by average value. The dashed line indicates the global average across all categories.}
    \label{fig:category_means_errorbars}
\end{figure*}

\begin{figure*}
    \centering
    \includegraphics[width=0.65\linewidth]{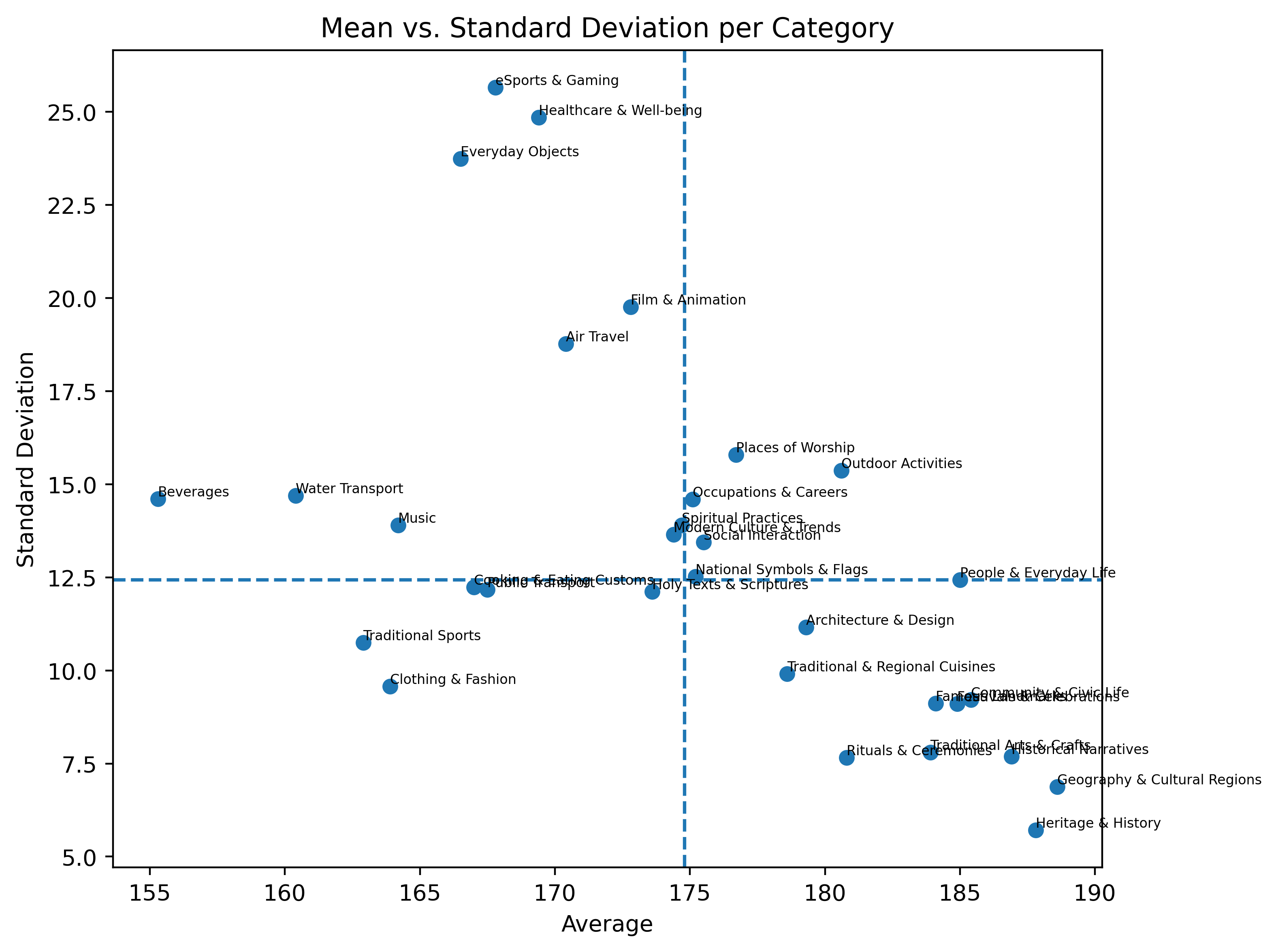}
    \vspace{-0.4cm}
    \caption{Scatter plot of mean score vs. standard deviation for each category. The vertical dashed line indicates the global average score (174.8), and the horizontal dashed line indicates the median variability across categories.
    }
    \label{fig:mean_std_scatter}
\end{figure*}

In Table \ref{tab:ablation_topic_query}, we provide detailed statistics for topic and query generation. We first generated 10 topics per subcategory using GPT-4o (See prompt in Listing \ref{lst:prompt_seed_topic_generation}), resulting in 5,580 topics in total. After manually verifying country relevance, we retained 5,445 topics and removed 135 that were not aligned with the target countries. 
We provide the country name, category, subcategory and its all associated topics to generate queries (see Listings \ref{lst:prompt_query_generation_english} and \ref{lst:prompt_query_generation_arabic} for the prompts used).
We asked each LLM to generate 50 queries per subcategory based on the provided topics.
We then prompted GPT-4o to assess cultural relevance and assign a relevancy score from 1 to 100 for each query. We manually reviewed a sample of the queries and their scores to determine an optimal threshold for two purposes: \textit{(i)} filtering out less relevant queries, and \textit{(ii)} reducing the total number of queries. Finally, we selected only the queries with a relevancy score of $\geq$80/100, yielding 97,678 queries in total. To understand per-subcategory query coverage, we analyzed the subcategory-wise distribution of queries across countries, as shown in Figures \ref{fig:category_means_errorbars} and \ref{fig:mean_std_scatter}. From these figures, it is clear that some subcategories exhibit relatively higher coverage with low variance (e.g., \textit{Heritage \& History}), whereas others show lower coverage (e.g., \textit{Beverages}).

\begin{table*}[!ht]
\centering
\caption{
Statistics of the number of topics, filtered version after manual verification. Followed by query generation and filtering. Columns use ISO 2-letter country codes (DZ = Algeria, BH = Bahrain, EG = Egypt, IQ = Iraq, JO = Jordan, KW = Kuwait, LB = Lebanon, LY = Libya, MA = Morocco, OM = Oman, PS = Palestine, QA = Qatar, SA = Saudi Arabia, SD = Sudan, SY = Syria, TN = Tunisia, AE = United Arab Emirates, YE = Yemen). Rel. score: relevance score.}
\label{tab:ablation_topic_query}
\setlength{\tabcolsep}{2pt} 
\scalebox{0.7}{%
\begin{tabular}{@{}l*{19}{r}@{}}
\toprule
\textbf{Metric} & DZ & BH & EG & IQ & JO & KW & LB & LY & MA & OM & PS & QA & SA & SD & SY & TN & AE & YE & \textbf{Total} \\ \midrule
\# Topics            & 310 & 310 & 310 & 310 & 310 & 310 & 310 & 310 & 310 & 310 & 310 & 310 & 310 & 310 & 310 & 310 & 310 & 310 & \textbf{5,580} \\
Man. Verified        & 304 & 305 & 310 & 303 & 297 & 301 & 306 & 307 & 310 & 309 & 294 & 310 & 287 & 310 & 283 & 303 & 310 & 296 & \textbf{5,445}\\
\# Queries  & 6,139 & 6,157 & 6,104 & 6,154 & 6,100 & 6,144 & 5,769 & 6,156 & 6,150 & 6,150 & 6,139 & 6,151 & 6,126 & 6,138 & 6,095 & 6,159 & 6,147 & 6,148 & \textbf{110,126}  \\
Rel. $\geq 80$ & 5,405 & 5,347 & 5,531 & 5,465 & 5,483 & 5,356 & 5,167 & 5,519 & 5,509 & 5,654 & 5,387 & 4,875 & 5,459 & 5,521 & 5,476 & 5,448 & 5,521 & 5,555 & \textbf{97,678} \\ \midrule
\multicolumn{20}{c}{\textbf{Query Stats}} \\ \midrule
Max & 192 & 195 & 196 & 191 & 194 & 195 & 196 & 193 & 197 & 196 & 191 & 200 & 196 & 195 & 195 & 193 & 197 & 197 \\
Min & 139 & 140 & 152 & 146 & 146 & 120 & 107 & 150 & 147 & 161 & 146 & 75 & 140 & 153 & 144 & 138 & 125 & 140 \\
Avg & 174.4 & 172.5 & 178.4 & 176.3 & 176.9 & 172.8 & 166.7 & 178.0 & 177.7 & 182.4 & 173.8 & 157.3 & 176.1 & 178.1 & 176.6 & 175.7 & 178.1 & 179.2 \\
Std & 13.0 & 13.7 & 12.3 & 12.4 & 11.6 & 15.3 & 24.2 & 11.5 & 11.9 & 9.7 & 12.3 & 31.1 & 15.5 & 11.4 & 12.3 & 14.4 & 16.0 & 12.1 \\
\bottomrule
\end{tabular}
}
\end{table*}


\subsection{Semantic and Cognitive Profile for QA}
\label{sec-app:semanticNcogni}

We complement the three QA types with a \textbf{semantic} and \textbf{cognitive} profile of questions grounded in the image content. To obtain semantic type information, we selected two images per subcategory and manually wrote questions. For each image, annotators (including several authors of the paper) independently inspected the content and drafted candidate questions. We then clustered the questions by semantic similarity across annotators and prompted an LLM to assign free-form labels. The resulting clusters were assigned one or more of the following labels: \emph{Location \& place identification; Scene interpretation \& context; Architectural features \& functions; Cultural significance \& heritage; Traditional clothing \& attire; Tourism \& cultural activities; Event \& activity type; Objects, animals \& food recognition; National symbols \& identity; Visual attributes; Recreational activities \& facilities}. This follows prior VQA practice of organizing questions by \emph{semantic types} to enable targeted analysis \citep{Kafle2017TDIUC,Hudson2019GQA}. In parallel, each question received a \emph{cognitive focus} tag, either \emph{knowledge-based} (requiring external/world knowledge) or \emph{commonsense-based} (requiring everyday reasoning), which is in line with benchmarks that explicitly separate knowledge-intensive and commonsense reasoning \citep{Marino2019OKVQA,schwenk2022okvqa,Zellers2019VCR}. Our semantic labels also align with established vision domains widely used for scene/place, attributes, and landmarks \citep{Zhou2018Places,Patterson2012SUNAttr,Weyand2020GLDv2}, facilitating transfer and comparison.

\subsection{Annotation Agreement for QA}
\label{sec-app:agreement}

We adopt the $r^{*}_{wg(j)}$ index \citep{james1984estimating} to quantify agreement on ordinal Likert ratings. The index compares the \emph{observed variance} ($S_X^2$) in annotators' ratings to the \emph{maximum possible variance} ($\sigma^2_{\text{mv}}$) under complete disagreement for the given bounded scale:
\begin{equation}
r^{*}_{wg(j)} = 1 - \frac{S_X^2}{\sigma^2_{\text{mv}}}.
\end{equation}
For a 5-point Likert scale with lower and upper bounds $X_L=1$ and $X_U=5$, the maximum variance is
\begin{equation}
\sigma^2_{\text{mv}} = 0.5(X_U^2 + X_L^2) - \left[0.5(X_U + X_L)\right]^2 .
\end{equation}

In Table~\ref{tab:qa_annotation_quality}, we report annotation agreement based on the Likert scale values for both human- and LLM-based annotations across three QA types such as open-ended, MCQ, and true/false (T/F) with LLMs evaluated across 18 countries and humans across 14 countries. Unless noted, quality means (A.Q., Q.Q., R.C.I., R.P.F.) are on a 1–5 scale; the T/F answer quality (A.Q.) uses a 1–3 scale. After linear rescaling, LLM T/F A.Q. of 2.90–2.95 (1–3) corresponds to approximately 4.80–4.90 on a 1–5 scale, and human T/F A.Q. of $\sim$3.00 maps to $\sim$5.00 indicating higher aggrement for binary answers. 

MCQ remains consistently strong across all mean quality dimensions ($\approx$4.69–4.95 on a 1–5 scale), with open-ended questions scoring only marginally lower. 
rwg$^{*}$ score (0–1 scale) is uniformly high, $\sim$0.79–0.99 for LLMs and $\sim$0.68–0.99 for humans, with humans generally higher, especially on T/F. Overall, the high agreement scores indicate strong annotation consistency and support the quality of the QA pairs and rationales. The comparatively lower agreement for the \textit{second T/F question} on question quality primarily reflects the intent of this item. It is designed to be \emph{false} and to reference content that is not present in the image. In early rounds, some annotators interpreted such questions as low quality 
and in some cases did not consistently apply the intended criterion. As discussed in Section~\ref{sec:introduction}, these questions are meant to measure hallucination behavior. After clarifying this guideline, subsequent rounds showed more consistent judgments.

\begin{table*}[ht]
\centering
\caption{
Annotation agreement scores for LLM and human annotations. The values for answer and question qualities, such as A.Q. and Q.Q., range between 1-5 except A.Q. for true/false 1-3. The values $r^*_{wg}$ scores ranges between 0-1. 
\textit{A.Q.} = Answer Quality Mean, 
\textit{A.Q. $r^*_{wg}$} = Answer Quality Inter-rater Agreement ($r^*_{wg}$), 
\textit{Q.Q.} = Question Quality Mean, 
\textit{Q.Q. $r^*_{wg}$} = Question Quality Inter-rater Agreement ($r^*_{wg}$),
\textit{R.C.I.} = Rationale Clarity \& Informativeness Mean, 
\textit{R.C.I. $r^*_{wg}$} = Rationale Clarity \& Informativeness Inter-rater Agreement ($r^*_{wg}$), 
\textit{R.P.F.} = Rationale Plausibility \& Faithfulness Mean, 
\textit{R.P.F. $r^*_{wg}$} = rationale plausibility \& faithfulness inter-rater agreement ($r^*_{wg}$).
}
\setlength{\tabcolsep}{3pt} 
\scalebox{0.75}{%
\begin{tabular}{lcccccccc}
\toprule
\textbf{QA Type} & \textbf{A.Q.} & \textbf{A.Q. $r^*_{wg}$} & \textbf{Q.Q.} & \textbf{Q.Q. $r^*_{wg}$} & \textbf{R.C.I.} & \textbf{R.C.I. $r^*_{wg}$} & \textbf{R.P.F.} & \textbf{R.P.F. $r^*_{wg}$} \\
\midrule
\multicolumn{9}{c}{\textbf{LLM-based Annotation (18 countries)}} \\ \midrule
Open-ended & 4.680 & 0.788 & 4.916 & 0.958 & 4.938 & 0.979 & 4.756 & 0.834 \\
MCQ         & 4.823 & 0.868 & 4.926 & 0.963 & 4.949 & 0.979 & 4.839 & 0.880 \\
T/F-0       & 2.898 & 0.963 & 4.947 & 0.973 & 4.967 & 0.986 & 4.874 & 0.902 \\
T/F-1       & 2.945 & 0.980 & 4.969 & 0.984 & 4.982 & 0.993 & 4.922 & 0.940 \\
\midrule
\multicolumn{9}{c}{\textbf{Human Annotation (14 countries)}} \\ \midrule
Open-ended & 4.746 & 0.950 & 4.785 & 0.959 & 4.687 & 0.944 & 4.675 & 0.945 \\
MCQ         & 4.771 & 0.957 & 4.804 & 0.966 & 4.692 & 0.951 & 4.724 & 0.952 \\
T/F-0       & 2.992 & 0.997 & 4.692 & 0.934 & 4.739 & 0.953 & 4.703 & 0.953 \\
T/F-1       & 2.994 & 0.998 & 3.704 & 0.676 & 4.727 & 0.952 & 4.693 & 0.952 \\
\bottomrule
\end{tabular}
}
\label{tab:qa_annotation_quality}
\end{table*}

\subsection{QA Audio}

\textbf{Audio Question Evaluation.}
We assessed audio quality using four standard metrics: \textit{(i)} Word Error Rate (WER) for transcription accuracy; %
\textit{(ii)} Speaker Cosine Similarity (SpkCos) for speaker consistency, based on embeddings from \texttt{spkrec-ecapa-voxceleb}~\citep{ravanelli2021speechbrain}; and \textit{(iii)} NISQA~\citep{mittag2021nisqa}, which predicts overall perceptual quality, including naturalness and distortion. 
For the transcription of both generated and human-recorded audio, we used Whisper-small for English~\cite{radford2023robust} and Fanar\footnote{A publicly accessible ASR API: \url{https://fanar.qa/}.} for Arabic.
\begin{table}[ht]
\centering
\caption{
Objective evaluation of generated (Gen.) and human-recorded (Human) audio for English and MSA. 
\textbf{WER} = Word Error Rate (lower is better), 
\textbf{SpkCos} = Speaker Cosine Similarity (higher is better), 
\textbf{NISQA} = Non-Intrusive Speech Quality Assessment (higher is better).
}
\label{tab:audio_quality_comparison}
\setlength{\tabcolsep}{1pt}
\scalebox{0.75}{%
\begin{tabular}{cccc|c}
\toprule
\textbf{Language} 
& \makecell{\textbf{WER} \\ \textbf{(Gen.)} $\downarrow$}
& \makecell{\textbf{SpkCos} \\ \textbf{(Gen.)} $\uparrow$}
& \makecell{\textbf{NISQA} \\ \textbf{(Gen.)} $\uparrow$}
& \makecell{\textbf{WER} \\ \textbf{(Human)} $\downarrow$} \\
\midrule
English & 6.17 & 0.58 & 4.33 & 11.86 \\
MSA     & 9.85 & 0.57 & 3.68 & 22.03 \\
\bottomrule
\end{tabular}
}
\end{table}

\subsection{Data Statistics}
\label{sec:descriptive_statistics}
\begin{figure}
    \centering
    \includegraphics[width=1.0\linewidth]{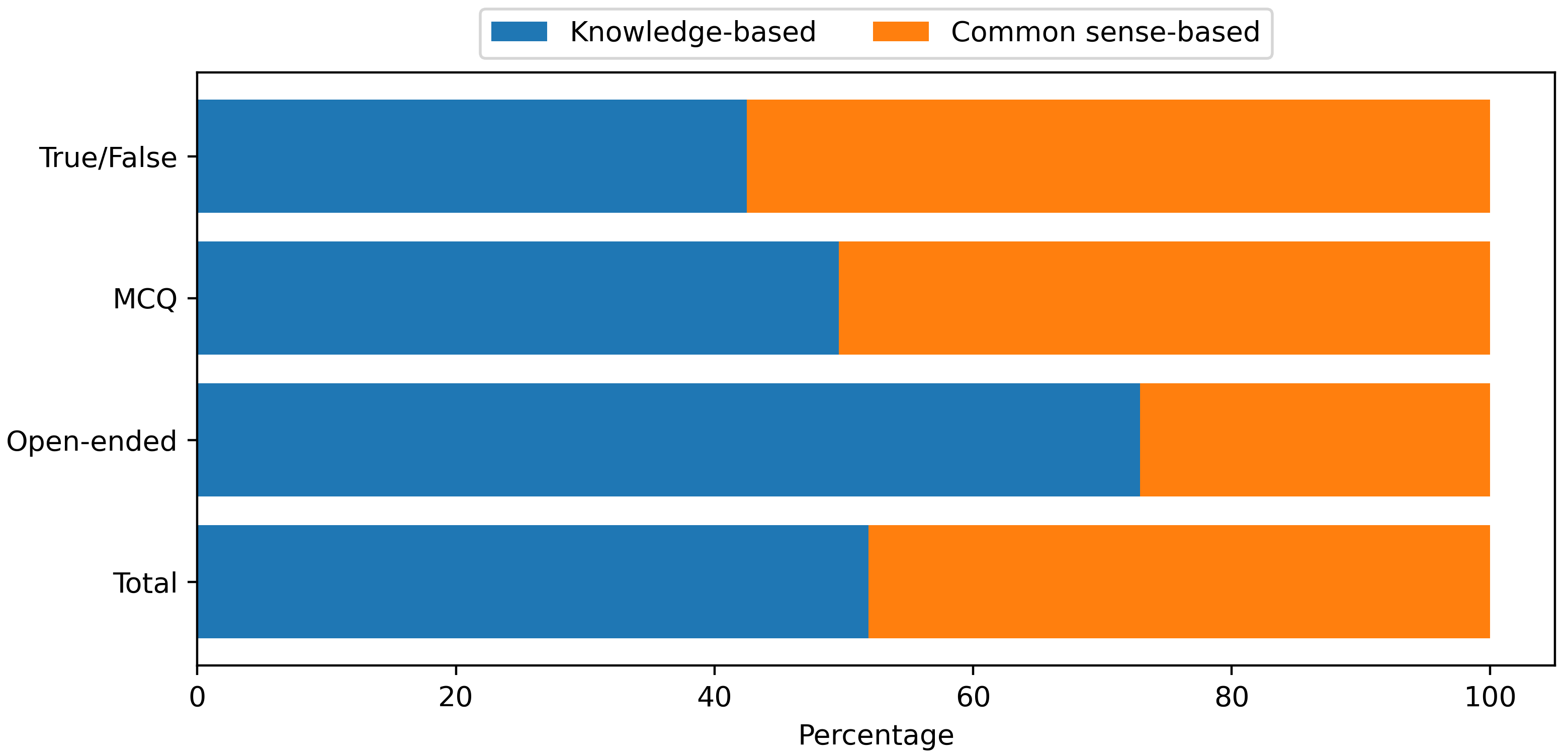}
    \caption{Distribution of commonsense and knowledge based for the whole dataset.}
    \label{fig:placeholder}
\end{figure}

Figure~\ref{fig:placeholder} shows the distribution between commonsense and knowledge-based questions.  Overall, the dataset is balanced (51.9\% knowledge-based vs.\ 48.1\% commonsense).  Open-ended questions are mostly knowledge-based (72.9\%), while true/false questions are mostly commonsense (57.5\%). Multiple-choice questions are almost evenly split, with 49.6\% being knowledge-based and 50.4\% being commonsense-based.

Table~\ref{tab:word_stats_mean_std} reports the mean and standard deviation of of text length (in words) across languages. In English, the statistics are consistently the highest across all cases, such as, descriptions (32.11$\pm$8.85), description reasons (37.84$\pm$5.59), questions (12.43$\pm$2.91), and rationales (19.56$\pm$4.28). Within the Arabic varieties, lengths decrease roughly as \textsc{MSA} $>$ Egyptian $\approx$ Levantine (e.g., description lengths: 26.66$\pm$7.63, 24.75$\pm$9.71, 24.45$\pm$11.75). Answers are short across languages (EN 7.71, MSA 6.79, Egy 6.89, Lev 6.53 words on average) yet show large variance (std $\approx$ 9--11), indicating a mix of one-word and phrase-level responses. Levantine exhibits the greatest dispersion (e.g., question std 6.79; rationale std 8.58), suggesting greater stylistic heterogeneity. Overall, the dataset provides substantive rationales ($\sim$16--20 words) and comparable question lengths across Arabic dialects, with English being more verbose.

In Tables \ref{tab:country_cat_tableA} and \ref{tab:country_cat_tableB}, we report the subcategory-wise data distribution across countries. Overall, the number of images per country ranges from 40K to 60K, although for a few countries it is lower, such as Qatar with 36K images.

\begin{table*}[ht]
\centering
\caption{Statistics of the image description, question, and rationals for language varieties. Numbers are represented as (mean$\pm$std).
}
\label{tab:word_stats_mean_std}
\setlength{\tabcolsep}{4pt} 
\scalebox{0.90}{%
\begin{tabular}{lrrrrr}
\toprule
\textbf{Lang} & \textbf{Description} & \textbf{Description Reason} & \textbf{Question} & \textbf{Answer} & \textbf{Rationale} \\
\midrule
EN  & 32.11 $\pm$ 8.85 & 37.84 $\pm$ 5.59 & 12.43 $\pm$ 2.91 & 7.71 $\pm$ 11.01 & 19.56 $\pm$ 4.28 \\
MSA & 26.66 $\pm$ 7.63 & 31.06 $\pm$ 5.42 & 9.95  $\pm$ 2.88 & 6.79 $\pm$ 9.68  & 16.90 $\pm$ 4.18 \\
Egy & 24.75 $\pm$ 9.71 & 28.73 $\pm$ 5.64 & 9.68  $\pm$ 3.98 & 6.89 $\pm$ 9.58  & 16.51 $\pm$ 5.55 \\
Lev & 24.45 $\pm$ 11.75& 28.62 $\pm$ 7.22 & 8.85  $\pm$ 6.79 & 6.53 $\pm$ 9.44  & 15.57 $\pm$ 8.58 \\
\bottomrule
\end{tabular}
}
\end{table*}

\paragraph{Recordings Statistics.}
In Table~\ref{tab:recordings_evaluation}, we report the number of MSA and English recording samples per country. Overall, the evaluation set supports cross-country, cross-variety analysis despite uneven per-country and per-language distributions. The speakers represent 12 countries (Table~\ref{tab:speakers_nationality}), diverse educational backgrounds from high school to PhD, and age groups ranging from 17 to 51 years. Hence, the diversity of recordings are varied by number of speakers, recording conditions, speaker demography, their dialectal accents. 
The English subset includes 28 speakers with an average duration of about 6 seconds per question, while the MSA subset includes 25 speakers with an average duration of about 5 seconds.

\begin{table*}[!tbh]
\centering
\caption{Number of MSA and English recording samples associated with open-ended and T/F questions per country used in evaluation. Country codes: DZ = Algeria, BH = Bahrain, EG = Egypt, IQ = Iraq, JO = Jordan, KW = Kuwait, LB = Lebanon, LY = Libya, MA = Morocco, OM = Oman, PS = Palestine, QA = Qatar, SA = Saudi Arabia, SD = Sudan, SY = Syria, TN = Tunisia, AE = United Arab Emirates, YE = Yemen.}
\label{tab:recordings_evaluation}
\setlength{\tabcolsep}{2pt} 
\scalebox{0.65}{%
\begin{tabular}{@{}l*{19}{c}@{}}
\toprule
& DZ & BH & EG & IQ & JO & KW & LB & LY & MA & OM & PS & QA & SA & SD & SY & TN & AE & YE & Total\\ \midrule
\textbf{\# of MSA} & 7,045 & 7,553 & 1,482 & 6,084 & 345  & 7,463 & 7,476 & 2,444 & 4,552 & 6,687 & 6,340 & 7,567 & 7,488 & 6,003 & 6,969 & 5,548 & 7,813 & 3,361  & 102,229 \\
\textbf{\# of English} & 4,549   & 6,014 & 12,090 & 5,372 & 6,053 & 7,070  & 2,112   & 5,955   & 4,539  & 4,273   & 4,323  & 7,526 & 1,270  & 4,270   & 4,595   & 4,345  & 4,609 & 3,275 & 92,240 \\ \midrule
\textbf{Total} 
& \textbf{11,594} & \textbf{13,567} & \textbf{13,572} & \textbf{11,456} & \textbf{6,398} & \textbf{14,533} & \textbf{9,588} & \textbf{8,399} & \textbf{9,091} & \textbf{10,960} & \textbf{10,663} & \textbf{15,093} & \textbf{8,758} & \textbf{10,273} & \textbf{11,564} & \textbf{9,893} & \textbf{12,422} & \textbf{6,636} & \textbf{194,469} \\
\bottomrule
\end{tabular}}
\end{table*}

\begin{table}[!tbh]
\centering
\caption{Distribution of speakers by nationality (English and MSA). Country codes use ISO 3166-1 alpha-2: DZ (Algeria), BD (Bangladesh), EG (Egypt), IN (India), PK (Pakistan), IQ (Iraq), PS (Palestine), SD (Sudan), SY (Syria), LY (Libya), MA (Morocco), TN (Tunisia).}
\label{tab:speakers_nationality}
\setlength{\tabcolsep}{3pt}
\scalebox{0.85}{%
\begin{tabular}{lcccccccccccc}
\toprule
\textbf{Lang.} & \textbf{DZ} & \textbf{BD} & \textbf{EG} & \textbf{IN} & \textbf{PK} & \textbf{IQ} & \textbf{PS} & \textbf{SD} & \textbf{SY} & \textbf{LY} & \textbf{MA} & \textbf{TN} \\
\midrule
\textbf{English} & 1 & 2 & 11 & 1 & 3 & 1 & 2 & 3 & 4 & - & - & - \\
\textbf{MSA}     & 1 & 1 & 12 & - & - & 1 & 1 & 6 & - & 1 & 1 & 1 \\
\bottomrule
\end{tabular}
}
\end{table}

\begin{table*}[t]
\centering
\caption{\textbf{Per-country counts (first half of categories)}. Abbreviations: Air = Air Travel; Arch = Architecture \& Design; Bev = Beverages; Cloth = Clothing \& Fashion; Civic = Community \& Civic Life; Cook = Cooking \& Eating Customs; Obj = Everyday Objects; Landm = Famous Landmarks; Fest = Festivals \& Celebrations; Film = Film \& Animation; Geo = Geography \& Cultural Regions; Health = Healthcare \& Well-being; Herit = Heritage \& History; HistN = Historical Narratives; Script = Holy Texts \& Scriptures; Modern = Modern Culture \& Trends.}
\label{tab:country_cat_tableA}
\scriptsize
\setlength{\tabcolsep}{3pt}
\resizebox{\linewidth}{!}{%
\begin{tabular}{l *{16}{S[table-format=4.0]}}
\toprule
\textbf{Country} & \textbf{Air} & \textbf{Arch} & \textbf{Bev} & \textbf{Cloth} & \textbf{Civic} & \textbf{Cook} & \textbf{Obj} & \textbf{Landm} & \textbf{Fest} & \textbf{Film} & \textbf{Geo} & \textbf{Health} & \textbf{Herit} & \textbf{HistN} & \textbf{Script} & \textbf{Modern} \\
\midrule
Algeria & 2135 & 1346 & 1596 & 2000 & 1815 & 1624 & 2632 & 1447 & 1352 & 1035 & 1741 & 1649 & 1288 & 1402 & 1065 & 1511 \\
Bahrain & 1778 & 1429 & 1231 & 2113 & 1452 & 1371 & 2059 & 1199 & 1245 & 1198 & 1528 & 1543 & 1159 & 1180 & 959 & 1033 \\
Egypt & 2384 & 1923 & 1739 & 2141 & 2025 & 1504 & 2666 & 1843 & 1589 & 1444 & 2596 & 1877 & 1546 & 1844 & 1126 & 1274 \\
Iraq & 1925 & 1257 & 1536 & 2317 & 2018 & 1365 & 2278 & 1151 & 1339 & 970 & 1966 & 1715 & 1294 & 1359 & 930 & 1517 \\
Jordan & 1864 & 1779 & 1899 & 2382 & 1664 & 1636 & 2400 & 1634 & 1612 & 1438 & 2367 & 1357 & 1949 & 1918 & 1290 & 1489 \\
Kuwait & 1700 & 1757 & 1173 & 2014 & 1363 & 1424 & 1961 & 1734 & 1107 & 895 & 1381 & 1707 & 1042 & 1092 & 879 & 1404 \\
Lebanon & 1941 & 1937 & 1780 & 1963 & 1952 & 1756 & 2094 & 1644 & 1315 & 1290 & 2183 & 1803 & 1730 & 1702 & 1405 & 1887 \\
Libya & 2078 & 1283 & 1662 & 2417 & 1622 & 1429 & 2678 & 1330 & 1201 & 1051 & 1736 & 1691 & 1067 & 1297 & 971 & 1518 \\
Morocco & 2248 & 1984 & 2130 & 2750 & 1850 & 2051 & 3010 & 2208 & 1714 & 1661 & 2700 & 1863 & 2102 & 2170 & 1354 & 1826 \\
Oman & 2064 & 2391 & 1418 & 1853 & 2092 & 1427 & 1963 & 1826 & 1342 & 1381 & 2169 & 1628 & 2080 & 1818 & 1260 & 1638 \\
Palestine & 2164 & 2028 & 1655 & 1716 & 1247 & 1820 & 2150 & 1682 & 1433 & 959 & 2725 & 1876 & 1597 & 1930 & 1456 & 1483 \\
Qatar & 1556 & 1948 & 886 & 1581 & 981 & 747 & 1378 & 2111 & 1331 & 670 & 1546 & 860 & 1149 & 1065 & 942 & 1202 \\
Saudi\_arabia & 2189 & 1536 & 1705 & 2504 & 1834 & 1889 & 2744 & 1872 & 1669 & 1380 & 2088 & 1360 & 1726 & 1675 & 1342 & 1863 \\
Sudan & 1811 & 1576 & 1558 & 1980 & 1651 & 1570 & 2598 & 1538 & 1417 & 874 & 1519 & 1595 & 1418 & 1405 & 1141 & 1249 \\
Syria & 1749 & 1407 & 1755 & 1995 & 1446 & 1670 & 2246 & 1460 & 1583 & 1268 & 1565 & 1730 & 1369 & 1536 & 1108 & 1352 \\
Tunisia & 1426 & 1831 & 2376 & 2418 & 1615 & 1907 & 2079 & 1738 & 1725 & 1267 & 1768 & 2167 & 1683 & 1601 & 1117 & 1410 \\
UAE & 2453 & 2701 & 1646 & 2277 & 1783 & 1734 & 2135 & 2361 & 1624 & 1332 & 2783 & 1706 & 1488 & 1961 & 1051 & 1972 \\
Yemen & 2142 & 1331 & 1389 & 2095 & 1857 & 1411 & 2269 & 1526 & 1444 & 1335 & 1660 & 1766 & 1421 & 1615 & 932 & 1818 \\
\bottomrule
\end{tabular}}
\end{table*}

\begin{table*}[t]
\centering
\caption{\textbf{Per-country counts (second half of categories)}. Abbreviations: Music = Music; Symbols = National Symbols \& Flags; Jobs = Occupations \& Careers; Outdoor = Outdoor Activities; People = People \& Everyday Life; Worship = Places of Worship; Transit = Public Transport; Rituals = Rituals \& Ceremonies; Social = Social Interaction; Spiritual = Spiritual Practices; Cuisine = Traditional \& Regional Cuisines; Crafts = Traditional Arts \& Crafts; Sports = Traditional Sports; Water = Water Transport; eSports = eSports \& Gaming.}
\label{tab:country_cat_tableB}
\scriptsize
\setlength{\tabcolsep}{3pt}
\resizebox{\linewidth}{!}{%
\begin{tabular}{l *{15}{S[table-format=4.0]} S[table-format=5.0]}
\toprule
\textbf{Country} & \textbf{Music} & \textbf{Symbols} & \textbf{Jobs} & \textbf{Outdoor} & \textbf{People} & \textbf{Worship} & \textbf{Transit} & \textbf{Rituals} & \textbf{Social} & \textbf{Spiritual} & \textbf{Cuisine} & \textbf{Crafts} & \textbf{Sports} & \textbf{Water} & \textbf{eSports} & \textbf{Total} \\
\midrule
Algeria & 1536 & 955 & 2057 & 1769 & 1936 & 1478 & 2063 & 1435 & 1675 & 1303 & 1644 & 1814 & 1565 & 2258 & 1252 & 50378 \\
Bahrain & 1594 & 1359 & 1392 & 1407 & 1397 & 1490 & 1545 & 1245 & 1158 & 1097 & 1296 & 1977 & 1592 & 1812 & 1455 & 44293 \\
Egypt & 1602 & 1415 & 1834 & 1812 & 1984 & 2043 & 2416 & 1272 & 1895 & 1216 & 1891 & 1639 & 1718 & 2250 & 1386 & 55894 \\
Iraq & 1481 & 1086 & 1673 & 1564 & 1699 & 1462 & 1854 & 1476 & 1753 & 1390 & 1443 & 1625 & 1781 & 1643 & 1610 & 48477 \\
Jordan & 1695 & 1782 & 1579 & 1648 & 2002 & 1424 & 1965 & 1385 & 1708 & 1366 & 1575 & 1844 & 2203 & 2023 & 1659 & 54536 \\
Kuwait & 1706 & 1417 & 1329 & 1982 & 1527 & 1311 & 1469 & 912 & 1233 & 928 & 1207 & 1686 & 1533 & 1780 & 1339 & 43992 \\
Lebanon & 1586 & 1431 & 1712 & 1170 & 2022 & 1657 & 2329 & 1606 & 1578 & 1830 & 1855 & 1760 & 2004 & 1592 & 1325 & 53839 \\
Libya & 1787 & 1076 & 1687 & 1722 & 1766 & 943 & 2104 & 1385 & 1463 & 1091 & 1534 & 1677 & 1835 & 1917 & 1437 & 48455 \\
Morocco & 1867 & 1862 & 1826 & 1785 & 1960 & 1643 & 2526 & 1603 & 2020 & 1501 & 2266 & 3039 & 2194 & 2338 & 1677 & 63728 \\
Oman & 1795 & 1410 & 1709 & 1971 & 1652 & 1816 & 2123 & 1495 & 1527 & 1326 & 1548 & 1813 & 1752 & 1724 & 1793 & 53804 \\
Palestine & 1877 & 1560 & 1922 & 1941 & 1629 & 1851 & 2285 & 1643 & 1373 & 1785 & 1557 & 1445 & 2339 & 2279 & 1548 & 54955 \\
Qatar & 854 & 1524 & 897 & 1129 & 929 & 1555 & 1224 & 945 & 904 & 918 & 786 & 1237 & 995 & 1395 & 826 & 36071 \\
Saudi Arabia & 1687 & 1392 & 1912 & 2063 & 2188 & 1931 & 1548 & 1842 & 1664 & 1870 & 1698 & 1781 & 2168 & 1759 & 1427 & 56306 \\
Sudan & 1598 & 959 & 1761 & 1981 & 1865 & 1612 & 2157 & 1043 & 1639 & 1330 & 1449 & 1722 & 1953 & 1970 & 1623 & 49562 \\
Syria & 1687 & 1077 & 1420 & 1889 & 1796 & 1347 & 2004 & 1293 & 1738 & 1396 & 1327 & 1501 & 2037 & 2227 & 1583 & 49561 \\
Tunisia & 1702 & 1445 & 1817 & 1844 & 1918 & 1841 & 1819 & 1519 & 1908 & 1519 & 1838 & 2020 & 2340 & 1783 & 1468 & 54909 \\
UAE & 1792 & 1684 & 1951 & 2082 & 2087 & 1830 & 2007 & 1634 & 1780 & 1853 & 1604 & 2018 & 2084 & 2118 & 1517 & 59048 \\
Yemen & 1964 & 1234 & 1915 & 1690 & 1550 & 1345 & 2335 & 1542 & 1676 & 1238 & 1620 & 1756 & 2236 & 1887 & 1678 & 51677 \\
\bottomrule
\end{tabular}}

\end{table*}

\subsection{Machine Translation Scores}
\label{apd:mt-results}
In Table~\ref{tab:dialectal-bleu-scores}, we compare BLEU for direct English$\rightarrow$Dialect (EN$\rightarrow$DIA) \textit{vs.} a two-step English$\rightarrow$MSA$\rightarrow$Dialect (EN$\rightarrow$MSA$\rightarrow$DIA) pipeline. For Levantine, the direct approach consistently outperforms the two-step pipeline across all test sets (avg.\ 13.81 vs.\ 7.74 BLEU). For Egyptian, results are mixed. The pipeline excels on \textit{arzen} and \textit{madar test nil 0 eg} (22.86 and 31.33 BLEU, respectively), while the direct model leads on \textit{madar test nil 1/2 eg}. These trends suggest that intermediate MSA can help for certain Egyptian benchmarks, whereas direct transfer is more reliable for Levantine.

\begin{table*}[!tbh]
\centering
\caption{Comparison of GPT-4.1 performance (BLEU score) with the 2-step pipeline performance. The EN->Dia columns correspond to GPT-4.1 performance while the EN->MSA->Dia column correspond to the 2-step pipeline performance.}
\setlength{\tabcolsep}{2pt} 
\scalebox{0.8}{
\begin{tabular}{@{}lrrlrr@{}}
\toprule
\multicolumn{3}{c}{\textbf{Levantine}}      & \multicolumn{3}{c}{\textbf{Egyptian}}        \\ \midrule
\textbf{Test Set} & \multicolumn{1}{l}{\textbf{EN->MSA-> DIA}} & \multicolumn{1}{l}{\textbf{EN --> DIA}} & \textbf{Test Set} & \multicolumn{1}{l}{\textbf{EN->MSA-> DIA}} & \multicolumn{1}{l}{\textbf{EN --> DIA}} \\ \midrule
madar test lev 0    & 9.54 & 20.25 & madar test nil 2 eg & 8.99  & 19.68 \\
madar test lev 0 lb & 9.03 & 12.10 & arzen               & 22.86 & 6.85  \\
LDC test            & 4.12 & 6.11  & madar test nil 1 eg & 7.91  & 18.63 \\
madar test lev 1 jo & 8.27 & 16.78 & madar test nil 0 eg & 31.33 & 18.12 \\ \midrule
\textbf{Average}             & \textbf{7.74} & \textbf{13.81} & \textbf{Average}             & \textbf{17.77} & \textbf{15.82} \\ \bottomrule
\end{tabular}}
\label{tab:dialectal-bleu-scores}
\end{table*}

\subsection{Data Split}
In Table~\ref{tab:data_split_by_country}, we report the data-split distribution for all countries. We used stratified sampling with subcategory as the label for the stratification. The dev and test splits each comprise $\sim$3.76\% of the data, yielding about 2K QAs per country per split.

\begin{table*}[ht]
\centering
\caption{Distribution of train, dev, and test splits across countries (two-letter ISO codes). DZ=Algeria, BH=Bahrain, EG=Egypt, IQ=Iraq, JO=Jordan, KW=Kuwait, LB=Lebanon, LY=Libya, MA=Morocco, OM=Oman, PS=Palestine, QA=Qatar, SA=Saudi Arabia, SD=Sudan, SY=Syria, TN=Tunisia, AE=United Arab Emirates, YE=Yemen.}
\setlength{\tabcolsep}{2pt} 
\scalebox{0.6}{%
\begin{tabular}{lrrrrrrrrrrrrrrrrrrrr}
\toprule
\textbf{Split} & \textbf{DZ} & \textbf{BH} & \textbf{EG} & \textbf{IQ} & \textbf{JO} & \textbf{KW} & \textbf{LB} & \textbf{LY} & \textbf{MA} & \textbf{OM} & \textbf{PS} & \textbf{QA} & \textbf{SA} & \textbf{SD} & \textbf{SY} & \textbf{TN} & \textbf{AE} & \textbf{YE} & \textbf{Total} \\
\midrule
\textbf{Train} & 46,503 & 40,414 & 51,982 & 44,594 & 50,658 & 40,123 & 49,962 & 44,576 & 59,843 & 49,925 & 51,101 & 32,178 & 52,419 & 45,680 & 45,683 & 51,020 & 55,166 & 47,805 & 859,632 \\
\textbf{Dev}   & 1,931  & 1,938  & 1,964  & 1,938  & 1,938  & 1,934  & 1,944  & 1,938  & 1,945  & 1,949  & 1,913  & 1,950  & 1,943  & 1,941  & 1,939  & 1,947  & 1,935  & 1,936  & 34,923 \\
\textbf{Test}  & 1,944  & 1,941  & 1,948  & 1,945  & 1,940  & 1,935  & 1,933  & 1,941  & 1,940  & 1,930  & 1,941  & 1,943  & 1,944  & 1,941  & 1,939  & 1,942  & 1,947  & 1,936  & 34,930 \\\midrule
\textbf{Total} & 50,378 & 44,293 & 55,894 & 48,477 & 54,536 & 44,992 & 53,839 & 48,455 & 63,728 & 53,804 & 54,955 & 36,071 & 56,306 & 49,562 & 49,561 & 54,909 & 59,048 & 51,677 & 929,485 \\
\bottomrule
\end{tabular}
}

\label{tab:data_split_by_country}
\end{table*}

\section{Experimental Setup}
\label{sec_app_exp_setup}
In Table \ref{tab:exp_modal_combo}, we report the details of the experimental setups for this study, which reports the number of models, modality and language varieties we have experimented with.

\begin{table*}[t]
\centering
\caption{Evaluated models across input modalities: text ($T$), text+image ($T{+}I$), speech ($S$), and speech+image ($S{+}I$), with text as output. A \checkmark\ indicates experiments conducted; $\times$ indicates not applicable. \textbf{Egy} = Egyptian Arabic (arz), \textbf{Lev} = Levantine Arabic (ajp), $Tr$ = transcription, $Tr{+}I$=transcription+image. FT = Fine-tuning.}

\setlength{\tabcolsep}{2pt} 
\scalebox{0.75}{%
\begin{tabular}{lcccccccccccccccc}
\toprule
\multirow{2}{*}{\textbf{Models}} &
\multicolumn{6}{c}{\textbf{English}} &
\multicolumn{6}{c}{\textbf{MSA}} &
\multicolumn{2}{c}{\textbf{Egy}} &
\multicolumn{2}{c}{\textbf{Lev}} \\
\cmidrule(lr){2-7}\cmidrule(lr){8-13}\cmidrule(lr){14-15}\cmidrule(lr){16-17}
& \textbf{T} & \textbf{S} & \textbf{T+I} & \textbf{Tr} & \textbf{Tr+I} & \textbf{S+I}
& \textbf{T} & \textbf{S} & \textbf{T+I} & \textbf{Tr} & \textbf{Tr+I} & \textbf{S+I}
& \textbf{T} & \textbf{T+I}
& \textbf{T} & \textbf{T+I} \\
\midrule
Gemini-pro [T,S,I]      & \checkmark & \checkmark & \checkmark & \checkmark & \checkmark & \checkmark & \checkmark & \checkmark & \checkmark & \checkmark & \checkmark & \checkmark & \checkmark & \checkmark & \checkmark & \checkmark \\
GPT-4.1 [T,I]           & \checkmark & $\times$ & \checkmark & \checkmark & \checkmark & $\times$ & \checkmark & $\times$ & \checkmark & \checkmark & \checkmark & \checkmark & \checkmark & \checkmark & \checkmark & \checkmark \\
GPT-4o-audio [S]        & $\times$ & \checkmark & $\times$ & $\times$ & $\times$ & $\times$ & $\times$ & \checkmark & $\times$ & $\times$ & $\times$ & $\times$ & $\times$ & $\times$ & $\times$ & $\times$ \\
GPT-5 [T,I]             & \checkmark & $\times$ & \checkmark & \checkmark & \checkmark & $\times$ & \checkmark & $\times$ & \checkmark & \checkmark & \checkmark & $\times$ & \checkmark & \checkmark & \checkmark & \checkmark \\
Phi-4 [T,S,I]           & \checkmark & \checkmark & \checkmark & \checkmark & \checkmark & \checkmark & \checkmark & \checkmark & \checkmark & \checkmark & \checkmark & \checkmark & \checkmark & \checkmark & \checkmark & \checkmark \\
Qwen-2.5 3B [T,S,I]     & \checkmark & \checkmark & \checkmark & \checkmark & \checkmark & \checkmark & \checkmark & \checkmark & \checkmark & \checkmark & \checkmark & \checkmark & \checkmark & \checkmark & \checkmark & \checkmark \\
Qwen-2.5 7B [T,S,I]     & \checkmark & \checkmark & \checkmark & \checkmark & \checkmark & \checkmark & \checkmark & \checkmark & \checkmark & \checkmark & \checkmark & \checkmark & \checkmark & \checkmark & \checkmark & \checkmark \\
FT--Qwen-2.5 3B [T,S,I] & \checkmark & \checkmark & \checkmark & \checkmark & \checkmark & \checkmark & \checkmark & \checkmark & \checkmark & \checkmark & \checkmark & \checkmark & \checkmark & \checkmark & \checkmark & \checkmark \\
\bottomrule
\end{tabular}
}\label{tab:exp_modal_combo}
\end{table*}

\section{Prompts}
\label{sec-app-prompts}

\subsection{Prompt for Topic \& Query Generation}
\label{sec-app-prompt-topic}

In Listing \ref{lst:prompt_seed_topic_generation}, we provide the prompts for generating seed topics. We provide the prompts for generating queries for English, MSA, and regional dialects
in Listings \ref{lst:prompt_query_generation_english}, and \ref{lst:prompt_query_generation_arabic}
Finally, we provide the prompt to generate cultural-relevance scores for seed queries in Listing \ref{lst:prompt_relevance_score}.

\begin{lstlisting}[language=TeX,caption={Prompt for generating 10 seed topics for each \textit{(country, category)} pair.},label={lst:prompt_seed_topic_generation}]

You are an AI specialized in generating highly relevant topics for image searches based on a given country, category, and subcategory. Your task is to generate a list of topics that are highly visual and well-suited for image searches. Ensure that the topics reflect the cultural, historical, or modern significance of the specified location.

Guidelines:
1. Topics should be engaging, highly visual, and unique to the specified country.
2. Ensure a mix of historical, modern, and futuristic aspects based on the subcategory.
3. Use well-known landmarks, cultural elements, or emerging trends where relevant.
4. Prioritize topics that are frequently searched for in image search engines.
5. If the subcategory is broad, ensure a diverse selection covering different aspects.
6. Do not include generic topics that could apply to any country; make them location-specific.
7. If the subcategory is too narrow and lacks visual topics, expand the scope slightly to include related themes.
8. Generate exactly 10 topics per request. If necessary, include related visual aspects.
9. Avoid redundant or overly generic suggestions.
10. Ensure diversity in the topics; avoid generating closely related topics.

JSON Format:
- Provide a list of topics, each being short, clear, and descriptive (e.g., 'Futuristic Skyscrapers of Dubai' or 'Traditional Wooden Temples of Japan').

    json
    [
        "Futuristic Skyscrapers",
        "Traditional Mosque"
    ]


Generate **exactly** 10 highly relevant topics for image search based on the following:

- Country: {country}
- Category: {category}
- Subcategory: {subcategory}

If there are fewer than 10 highly relevant topics, expand the scope slightly to related visual themes. 

Ensure the topics are visually engaging, related to the specified country, and match common image search behavior. 
The topics should cover a mix of historical, modern, and futuristic elements unique to the location.

\end{lstlisting}

\begin{lstlisting}[language=TeX,caption={Prompt to generate queries in English.},label={lst:prompt_query_generation_english}]
You are an expert at generating **highly relevant, human-like image search queries** optimized for **Google Image Search**.

Your task is to generate **50 unique search queries** that reflect **natural human behavior**, including:
- **Typos, slang, informal expressions**, and **incomplete or autocomplete-style phrases**.
- Use **descriptive visual terms** such as "HD," "4K," "wallpaper," "real photo," "aesthetic," "close-up," "latest pics," etc.
- Mimic **real-world search styles**, including:
- Pure **keywords**
- **Questions** (e.g., "what do [topic] look like")
- **Autocomplete-like fragments** (e.g., "best pics of...")
- **Trending styles** (e.g., "free download,")

Incorporate **localized and culturally relevant elements** from the country provided, including:
- Dialects, slang, and spelling variations
- Famous **cities, landmarks**, or **cultural symbols**
- Country-specific visual cues, aesthetics, or references

Queries should be:
- **Short, human-like and natural-sounding** (2-5 words on average)
- **Highly visual** and suitable for image search intent
- Focused on the **topics**, not just the country, category or subcategory
- Always returned in **strict JSON format**:
json
{
    "queries": [
        "query 1",
        "query 2",
        "... up to query 50"
    ]
}
    
Remember to generate exactly 50 unique queries, ensuring a diverse range of search styles and incorporating elements specific to the given country. Focus on creating queries that real users might type when searching for images related to the provided topic.

Generate **50** unique, human-like image search queries** based on below information:

- Country: {country}  
- Category: {category}  
- Subcategory: {subcategory}  
- Topics: {topic}

\end{lstlisting}

\begin{lstlisting}[language=TeX,caption={Prompt for generating queries in MSA and dialects.},label={lst:prompt_query_generation_arabic}]

You are an expert at generating **highly realistic, human-like image search queries in Arabic** optimized for **Google Image Search**.

Your task is to generate 50 unique Arabic image search queries based on a given **country**, **category**, **subcategory**, and **topic**. These queries should reflect **how real people from the Arab world search for images**, using both **Modern Standard Arabic (MSA)** and **country-specific dialects** where appropriate.

Follow these guidelines to generate the queries:

1. Reflect natural human behavior:
   - Use informal phrasing, spelling mistakes, colloquial expressions, and incomplete or autocomplete-style fragments.
   - Vary punctuation, phrasing, and structure - some queries should be formal, others casual or conversational.
   - Mimic how people write queries on their phones or in autocomplete (e.g., `\<أجمل صور المغرب>`, `\<وين ألقى صور الشوارع>`).

2. Use visual and search-specific descriptors in Arabic:
   - Words like: `\<صور>`, `\<خلفيات>`, `\<تنزيل>`, `\<تحميل مجاني>`, `\<أجمل صور>`, `4K`, `\<صور HD>`, `\<حقيقية>`, `\<من انستقرام>`, `\<خلفية جوال>`.

3. Mimic real-world Arabic search styles:
   - Pure keywords
   - **Keyword-based**: e.g., `\<صور اللبس المغربي>`, `\<خلفيات الصحراء الجزائرية>`
   - **Questions**: e.g., `\<أين أجد صور الأسواق القديمة في اليمن؟>`
   - **Incomplete phrases**: e.g., `\<أجمل صور من عمان>`, `\<تنزيل صور تقليدية>`

4. Use localized and culturally relevant language:
   - Mention famous places or cultural features (e.g., pyramids, mosques, old souks, traditional outfits).
   - Dialects to consider: Egyptian, Gulf, Levantine, Maghrebi - depending on the country.

Ensure the queries:
- Are **short (2-5 words)**, natural-sounding, and visually oriented
- **Focus heavily on the topic**
- Avoid any overly formal, robotic phrasing

Return the results in the following **strict JSON format only**:
json
{
    "queries": [
        "query 1 in Arabic",
        "query 2 in Arabic",
        "... up to query 50 in Arabic"
    ]
}

Generate **50** unique, human-like image search queries in Arabic based on below information:

- Country: {country}
- Category: {category}
- Subcategory: {subcategory}
- Topics: {topic}

\end{lstlisting}

\begin{lstlisting}[language=TeX,caption={Prompt for generating cultural-relevance score of search queries. The place-holder represents the list of queries.},label={lst:prompt_relevance_score}]
You are an expert in evaluating search query effectiveness for image search. Your task is to rank image search queries based on their relevance to a given location. Focus on specificity, uniqueness, and cultural significance when ranking them. Assign a relevance score from 1 to 100 and return the output in JSON format.

Given the following list of **image search queries** related to {location}, **evaluate and assign a relevance score to every single query**.

Each query must receive a **relevance score from 1 to 100**, where:
- 100 represents the highest relevance.
- Higher scores go to queries highlighting **iconic landmarks, cultural elements, or unique aspects of {location}**.
- Queries mentioning a **location outside of {location}** should receive a **low relevance score**.
- **DO NOT skip any query** - every query in the list must be assigned a score.
- Queries are in **Arabic and English** - evaluate both equally.
- Queries those are not related to {location} should receive a very low score.

### **List of Queries:**  
json
{json.dumps(query_list, ensure_ascii=False)}


**Expected JSON Output Format:**        
json
[
    {{"Q": "Eiffel Tower at sunset", "score": 100}},
    {{"Q": "Paris street art", "score": 90}},
]
\end{lstlisting}

\subsection{Prompt for Image Description Generation}
\label{sec-prompt-img-desc-gen}
We provide the prompt that we used for image description generation and image categorization in Listing \ref{lst:prompt_generate_image_description}.

\begin{lstlisting}[language=TeX,caption={Prompt for generating bilingual image description and categorization.},label={lst:prompt_generate_image_description}]
You are an AI assistant specializing in image analysis, filtering, and categorization for question-answering (QA) systems. Your task is to **describe, classify, and assess** images based on their relevance and suitability.
### **1. Image Description**
- Provide a **concise, objective** description of the image.
- Extract readable text (if any) and include it under "extracted_text".

### **2. Image Categorization**
Classify the image into **one of the following categories**:
- **Photograph** - Real-world photo.
- **Illustration** - Hand-drawn or digital artwork (e.g., sketches, comics). Excludes branded mascots in ads.
- **Advertisement** - Promotional content with branding, pricing, slogans, or call-to-action text. Includes banners, flyers, or sponsored content.
- **Screenshot/UI Capture** - Software, websites, or apps.
- **Meme/Text Overlay** - An image with overlaid text, often humorous or social.
- **Chart/Infographic** - A diagram or data visualization, such as an infographic or graph.
- **Other** - Any content that does not fit the above categories.

### **3. Suitability Assessment**
- Determine if the image is **suitable** for a QA system based on:
- **Clarity** (clear, readable, and interpretable).
- **Relevance** (must align with the user-provided **topic** and **subtopic**).
- **Content** (must not contain inappropriate elements).
- Provide a **justification** for the suitability decision.

### **4. Response Format (JSON)**
Return results in **structured JSON format**:
{{
    "description": "<concise image description>",
    "extracted_text": "<text extracted from image (if any)>",
    "image_category": "<category>",
    "status": "<suitable/not_suitable>",
    "reason": "<brief explanation>"
}}


Analyze the given image in the context of **Topic: {category.lower()}** and **Subtopic: {subcategory.lower()}**.

**Image:** {image}

- **Describe the image.**  
- **Extract readable text (if any).**  
- **Classify the image into a predefined category.**  
- **Assess if it is suitable** for a QA system based on clarity, relevance, and content.  
        
\end{lstlisting}

\subsection{Prompt for Generating Question-Answer}
\label{sec-prompt-qa-gen}
We provide a prompt for generating four cultural question–answer pairs per image, as shown in Listing \ref{lst:prompt_generate_image_description}.

\begin{lstlisting}[language=TeX,caption={Prompt for generating four cultural question-answer pairs per image.},label={lst:prompt_generating_qa}]
You are an AI assistant specializing in Visual Question Answering (VQA). Your task is to analyze the given image and generate high-quality Question-Answer (Q&A) pairs for benchmarking and training large language models (LLMs).

Follow these guidelines carefully:

1. Types of Q&A Pairs (generate all for each image):
    1. Open-ended: A free-form question with an informative answer based on the image.
    2. Multiple-choice: A question with three plausible options, clearly marking the correct answer.
    3. True/False: A question-answer pair that can be answered with 'True' or 'False'. 

    For type 1 and 2 you should generate one QA pair for each. For type 3 you should generate two QA pairs, one with True and one with False. 


2. Semantic Focus:
    - Use the following semantic labels to guide your questions. Match the image content to the most relevant labels:
        - Location and Place Identification
        - Scene Interpretation and Context
        - Architectural Features and Functions
        - Cultural Significance and Heritage
        - Traditional Clothing and Attire
        - Tourism and Cultural Activities
        - Event and Activity Type
        - Objects, Animals, and Food Recognition
        - National Symbols and Identity
        - Visual Attributes
        - Recreational Activities and Facilities

3. Cognitive Focus:
    - Ensure a balanced mix of:
        - Knowledge-based questions (requiring factual knowledge related to the image).
        - Common sense-based questions (requiring general reasoning or everyday knowledge to answer).
    - Assign a label to each question indicating its cognitive focus (knowledge-based or common sense-based).

4. Language:
    - All Q&A pairs must be written in native-sounding English.

5. Question Quality:
    - Ensure the questions are natural, conversational, and human-like.
    - Vary the phrasing and difficulty across the different question types. Questions should be engaging and thought-provoking. A mix of simple and complex questions is encouraged.

6. Answer Quality:
    - Answers must be factually correct, clear, concise, and well-structured.
    - Use correct grammar and maintain high readability.

7. Cultural Sensitivity:
    - Avoid stereotypes or cultural misrepresentations.
    - Ensure cultural references are accurate and specific to the image.

8. Context Utilization:
    - Use the provided image description, category, and subcategory to enrich the context while formulating the questions.

9. Reasoning:
    - For each Q&A pair, also provide a short explanation justifying why the answer is correct. Limit the explanation to less than 100 words.

Strictly follow these instructions to ensure the generated VQA data is of the highest quality and suitable for model evaluation and fine-tuning.

### **Output Format (JSON):**  
json
{{
    "open-ended": [
        {{"question_en": "...", "answer_en": "...", "rationale": "...", "cognitive_focus": "...","semantic_focus": ["...","..."]}},                
    ],
    "multiple-choice": [
        {{"question_en": "...", "options_en": ["..."], "correct_answer_en": "...", "rationale": "...","cognitive_focus": "...","semantic_focus": ["...","..."]}},
    ],
    "True/False": [
        {{"question_en": "...", "answer_en": "...", "rationale": "...,"cognitive_focus": "..."","semantic_focus": ["...","..."]}},
         {{"question_en": "...", "answer_en": "...", "rationale": "...,"cognitive_focus": "..."","semantic_focus": ["...","..."]}},
    ]
}}


Analyze the given image and generate **question-answer pairs with their rationales for each type: 1) Open-ended, 2) Multiple-choice, 3. True/False QA pairs**.  

**Image:** {image}

Use the following information as an additional context for generating questions:
**Description:** {description}
**Category:** {category}
**Subcategory:** {subcategory}

\end{lstlisting}

\section{Details of the Results}
\label{sec-app: details_rslt}

We report text-only and text+image results on the full dataset in Table~\ref{tab:msa_full}, comparing Gemini-Pro, Qwen-2.5 (Omni-3B), and the fine-tuned Qwen-2.5 (Omni-3B). Figure~\ref{fig:judge_merged} summarizes LLM-as-a-judge scores across modalities and models. Table~\ref{tab:speech_combination_modal_results} further reports results across models, modalities, and languages for all question types.

\begin{table*}[t]
\centering
\caption{\textbf{Open-ended} LLM-as-Judge results (English on top, MSA below) on the full dataset. \textbf{Country codes} are shown as columns: DZ (Algeria), BH (Bahrain), EG (Egypt), IQ (Iraq), JO (Jordan), KW (Kuwait), LB (Lebanon), LY (Libya), MA (Morocco), OM (Oman), PS (Palestine), QA (Qatar), SA (Saudi Arabia), SD (Sudan), SY (Syria), TN (Tunisia), AE (UAE), YE (Yemen). 
\textbf{Model settings}: T = Text, T+I = Text+Image.}
\label{tab:msa_full}
\setlength{\tabcolsep}{2.7pt} %
\scalebox{0.75}{
\begin{tabular}{@{}l*{19}{c}@{}}
\toprule
 & DZ & BH & EG & IQ & JO & KW & LB & LY & MA & OM & PS & QA & SA & SD & SY & TN & AE & YE & \textbf{Avg.}\\
\midrule
\multicolumn{20}{c}{\textbf{English}} \\ \midrule

Gemini-Pro (T) & 5.60 & 5.78 & 5.78 & 5.40 & 5.74 & 5.64 & 5.62 & 5.53 & 5.74 & 5.85 & 5.63 & 5.78 & 5.88 & 5.48 & 5.61 & 5.58 & 6.04 & 5.68 & \textbf{5.69}\\
Gemini-Pro (T+I) & 7.00 & 7.01 & 6.61 & 6.80 & 6.73 & 6.97 & 7.00 & 6.91 & 6.96 & 6.96 & 6.82 & 6.90 & 6.80 & 6.90 & 6.94 & 6.96 & 6.98 & 6.98 & \textbf{6.90}\\
Qwen2.5 (T) & 3.74 & 3.81 & 3.96 & 3.89 & 3.86 & 3.79 & 3.84 & 3.77 & 3.86 & 3.98 & 3.92 & 3.88 & 3.87 & 3.84 & 3.83 & 3.71 & 3.90 & 3.77 & \textbf{3.84}\\
Qwen2.5 (T+I) & 4.86 & 4.91 & 4.92 & 5.00 & 4.84 & 4.90 & 4.89 & 4.83 & 4.95 & 5.00 & 5.03 & 4.95 & 4.99 & 4.94 & 4.78 & 4.74 & 4.97 & 4.90 & \textbf{4.91}\\
Qwen2.5-FT (T) & 5.78 & 5.89 & 5.88 & 5.72 & 5.89 & 5.63 & 5.71 & 5.76 & 6.07 & 6.10 & 5.81 & 6.05 & 5.92 & 5.66 & 5.82 & 5.79 & 6.07 & 5.75 & \textbf{5.85}\\
Qwen2.5-FT (T+I) & 7.48 & 7.41 & 7.50 & 7.43 & 7.53 & 7.42 & 7.45 & 7.46 & 7.58 & 7.63 & 7.51 & 7.46 & 7.51 & 7.37 & 7.44 & 7.40 & 7.54 & 7.37 & \textbf{7.47}\\ \midrule
\multicolumn{20}{c}{\textbf{MSA}} \\\midrule
Gemini-Pro (T) & 5.44 & 5.59 & 5.55 & 5.28 & 5.54 & 5.37 & 5.50 & 5.42 & 5.53 & 5.65 & 5.46 & 5.56 & 5.65 & 5.35 & 5.45 & 5.37 & 5.69 & 5.53 & \textbf{5.50}\\
Gemini-Pro (T+I) & 7.73 & 7.20 & 6.84 & 7.00 & 6.95 & 7.23 & 7.25 & 7.19 & 7.12 & 7.20 & 7.04 & 7.01 & 6.94 & 7.14 & 7.15 & 7.23 & 7.17 & 7.20 & \textbf{7.14}\\
Qwen2.5 (T) & 4.70 & 4.82 & 4.94 & 4.74 & 4.94 & 4.58 & 4.80 & 4.71 & 4.77 & 4.87 & 4.90 & 4.78 & 4.82 & 4.73 & 4.70 & 4.71 & 4.79 & 4.74 & \textbf{4.78}\\
Qwen2.5 (T+I) & 5.16 & 5.30 & 5.39 & 5.27 & 5.41 & 5.14 & 5.30 & 5.20 & 5.29 & 5.31 & 5.39 & 5.14 & 5.34 & 5.21 & 5.25 & 5.27 & 5.29 & 5.23 & \textbf{5.27}\\
Qwen2.5-FT (T) & 6.39 & 6.41 & 6.49 & 6.25 & 6.42 & 6.20 & 6.30 & 6.27 & 6.47 & 6.62 & 6.44 & 6.52 & 6.53 & 6.17 & 6.38 & 6.34 & 6.58 & 6.31 & \textbf{6.40}\\
Qwen2.5-FT (T+I) & 8.29 & 8.35 & 8.36 & 8.24 & 8.33 & 8.32 & 8.30 & 8.29 & 8.36 & 8.36 & 8.31 & 8.23 & 8.34 & 8.17 & 8.18 & 8.27 & 8.27 & 8.24 & \textbf{8.29}\\
\bottomrule
\end{tabular}}
\end{table*}

\begin{table*}[t]
\centering
\caption{Evaluation results across languages and modalities. 
\textbf{F1} = F1 BERTScore, 
\textbf{Judge} = LLM-as-judge score, 
\textbf{Acc} = accuracy, 
\textbf{T} = text, 
\textbf{T+I} = text+image. 
Gemini = Gemini-2.5-pro. 
Underlined values denote the best text-only performance for each dialect, while bold values indicate the best text+image performance for Open-Ended Judge, MCQ, and True/False.
}
\label{tab:results_text_image}
\vspace{-0.2cm}
\setlength{\tabcolsep}{2pt} 
\scalebox{0.6}{
\begin{tabular}{llccccc|ccccc}
\toprule
\multirow{2}{*}{\textbf{Model}} & \multirow{2}{*}{\textbf{Modality}} 
& \multicolumn{5}{c|}{\textbf{English}} & \multicolumn{5}{c}{\textbf{MSA}} \\
\cmidrule(lr){3-7} \cmidrule(lr){8-12}
 & & \textbf{OE (F1)} & \textbf{OE (Judge)} & \textbf{MCQ (Acc)}  & \textbf{TF1 (Acc)}  & \textbf{TF2 (Acc)} 
   & \textbf{OE (F1)} & \textbf{OE (Judge)} & \textbf{MCQ (Acc)} & \textbf{TF1 (Acc)} & \textbf{TF2 (Acc)} \\
\midrule
GPT-4.1 & T   & 0.60 & 6.26 & 0.82 & 0.63 & 0.77 & 0.58 & 6.36 & 0.75 & 0.69 & 0.66 \\
        & T+I & 0.73 & \textbf{8.60} & \textbf{0.98 }& 0.97 & \textbf{0.99} & 0.62 & \textbf{8.36} & 0.96 & 0.96 & 0.98 \\  
GPT-5   & T   & 0.61 & \underline{6.39} & 0.80 & 0.58 & 0.84 & 0.55 & \underline{6.39} & 0.74 & 0.78 & 0.56 \\
        & T+I & 0.66 & 8.46 & 0.98 & 0.97 & \textbf{0.99} & 0.57 & 8.10 & 0.97 & 0.97 & 0.98 \\
Gemini  & T   & 0.57 & 5.50 & 0.77 & 0.76 & 0.71 & 0.54 & 5.69 & \underline{0.74} & 0.71 & 0.74 \\
        & T+I & 0.63 & 7.14 & 0.97 & 0.93 & 0.98 & 0.56 & 6.90 & 0.96 & 0.94 & 0.98 \\
Qwen-7B & T   & 0.57 & 5.11 & 0.70 & 0.61 & 0.64 & 0.53 & 4.45 & 0.56 & 0.77 & 0.41 \\
        & T+I & 0.64 & 5.10 & 0.97 & \textbf{0.98} & 0.98 & 0.55 & 4.45 & 0.91 & 0.96 & 0.94 \\
Phi-4   & T   & 0.55 & 5.01 & 0.66 & 0.43 & 0.78 & 0.51 & 3.71 & 0.50 & 0.61 & 0.54 \\
        & T+I & 0.59 & 6.22 & 0.86 & 0.85 & 0.95 & 0.51 & 4.16 & 0.67 & 0.89 & 0.63 \\
Qwen-3B & T   & 0.54 & 4.78 & 0.67 & 0.54 & 0.73 & 0.52 & 3.84 & 0.48 & 0.83 & 0.36 \\
        & T+I & 0.50 & 5.27 & 0.35 & 0.97 & 0.97 & 0.52 & 4.91 & 0.39 & 0.96 & 0.87 \\ \hline
FT (Qwen-2.5-3B) & T & 0.73 & \underline{6.39} & \underline{0.92} & \underline{0.91} & \underline{0.88} & 0.64 & 5.85 & 0.89 & \underline{0.90} & \underline{0.86} \\
 & T+I & 0.78 & 8.29 & \textbf{0.98} & \textbf{0.98} & \textbf{0.99} & 0.67 & 7.47 & \textbf{0.97} & \textbf{0.97} & \textbf{0.98} \\        
\midrule
& 
& \multicolumn{5}{c|}{\textbf{Egyptian Arabic}} & \multicolumn{5}{c}{\textbf{Levantine Arabic}} \\
\midrule
GPT-4.1 & T   & 0.56 & 6.07 & 0.61 & 0.72 & 0.57 & 0.57 & \underline{6.41} & 0.71 & 0.74 & 0.55 \\
        & T+I & 0.62 & \textbf{8.30} & \textbf{0.81} & 0.95 & \textbf{0.98} & 0.62 & \textbf{8.39} & \textbf{0.92} & 0.96 & \textbf{0.98} \\
GPT-5   & T   & 0.53 & \underline{6.18} & 0.60 & 0.82 & 0.44 & 0.55 & 6.31 & 0.70 & 0.83 & 0.44 \\
        & T+I & 0.55 & 7.86 & \textbf{0.81} & 0.96 & 0.98 & 0.57 & 8.03 & \textbf{0.92} & 0.97 & \textbf{0.98} \\
Gemini  & T   & 0.51 & 5.42 & 0.58 & 0.66 & 0.75 & 0.52 & 5.60 & 0.70 & 0.70 & 0.74 \\
        & T+I & 0.52 & 6.62 & \textbf{0.81} & 0.93 & 0.97 & 0.55 & 6.85 & \textbf{0.92} & 0.94 & \textbf{0.98} \\
Qwen-7B & T   & 0.48 & 4.07 & 0.45 & 0.75 & 0.45 & 0.49 & 4.23 & 0.55 & 0.80 & 0.40 \\
        & T+I & 0.49 & 5.70 & 0.75 & 0.94 & 0.90 & 0.51 & 5.76 & 0.87 & 0.95 & 0.93 \\
Phi-4   & T   & 0.46 & 2.94 & 0.39 & 0.66 & 0.47 & 0.47 & 3.24 & 0.45 & 0.70 & 0.48 \\
        & T+I & 0.46 & 3.56 & 0.49 & 0.85 & 0.45 & 0.48 & 3.76 & 0.61 & 0.89 & 0.46 \\
Qwen-3B & T   & 0.47 & 3.21 & 0.36 & 0.81 & 0.33 & 0.49 & 3.60 & 0.44 & 0.82 & 0.37 \\
        & T+I & 0.45 & 4.18 & 0.27 & 0.95 & 0.75 & 0.46 & 4.32 & 0.32 & 0.96 & 0.81 \\ \hline
FT (Qwen-2.5-3B) & T & 0.64 & 5.85 & \underline{0.72} & \underline{0.89} & \underline{0.84} & 0.64 & 5.96 & \underline{0.84} & \underline{0.90} & \underline{0.85} \\
 & T+I & 0.67 & 7.35 & 0.79 & \textbf{0.97} & 0.97 & 0.67 & 7.56 & 0.91 & \textbf{0.97} & \textbf{0.98} \\       
\bottomrule
\end{tabular}}
\vspace{-0.3cm}
\end{table*}

\begin{table*}[h]
\centering
\caption{Evaluation results across languages and \textbf{speech modality} combinations. \textbf{F1} = F1 BERTScore, \textbf{Judge} = LLM-as-judge score (GPT-4.1), \textbf{Acc} = accuracy, \textbf{T} = text, \textbf{Tr} = transcription, \textbf{T+I} = text+image, \textbf{Tr+I} = transcription+image. Judge scores range from 1 to 10. Gemini = Gemini-2.5-pro. The best model across all modalities for Open-Ended Judge and True/False Accuracy, is shown in bold.
}
\label{tab:speech_combination_modal_results}
\scalebox{0.75}{
\begin{tabular}{ll rrrr rrrr}
\toprule
\multicolumn{1}{c}{} & \multicolumn{1}{c}{} & \multicolumn{4}{c}{\textbf{English}} & \multicolumn{4}{c}{\textbf{MSA}} \\ \midrule
\multicolumn{1}{c}{} & \multicolumn{1}{c}{} & \multicolumn{2}{c}{\textbf{Open-ended}} & \multicolumn{1}{c}{\textbf{TF 1}} & \multicolumn{1}{c}{\textbf{TF 2}} &
\multicolumn{2}{c}{\textbf{Open-ended}} & \multicolumn{1}{c}{\textbf{TF 1}} & \multicolumn{1}{c}{\textbf{TF 2}} \\ \midrule
\textbf{Model} & \textbf{Modality} & \textbf{F1} & \textbf{Judge} & \textbf{Acc} & \textbf{Acc} & \textbf{F1} & \textbf{Judge} & \textbf{Acc} & \textbf{Acc} \\
\midrule

GPT-4.1 & T & 0.60 & 6.16 & 0.51 & 0.84 & 0.58 & 6.35 & 0.69 & 0.66 \\
 & T+I & 0.75 & \textbf{8.80} & 0.97 & \textbf{0.99 }& 0.64 &\textbf{ 8.47} & 0.96 & \textbf{0.99} \\
 & Tr & 0.57 & 5.36 & 0.41 & 0.83 & 0.56 & 5.40 & 0.45 & 0.75 \\
 & Tr+I & 0.73 & 8.38 & 0.80 & 0.97 & 0.62 & 8.02 & 0.81 & 0.94 \\
\midrule

GPT-4o & S & 0.62 & 5.29 & 0.42 & 0.78 & 0.55 & 3.91 & 0.55 & 0.61 \\
\midrule

GPT-5 & T & 0.49 & 5.86 & 0.37 & 0.91 & 0.51 & 6.03 & 0.42 & 0.86 \\
 & T+I & 0.60 & 8.11 & 0.92 & \textbf{0.99} & 0.54 & 7.89 & 0.90 & \textbf{0.99} \\
 & Tr & 0.46 & 5.02 & 0.26 & 0.91 & 0.49 & 5.23 & 0.28 & 0.87 \\
 & Tr+I & 0.58 & 7.57 & 0.73 & 0.97 & 0.52 & 7.41 & 0.75 & 0.95 \\
\midrule

Qwen-2.5-7B & T & 0.56 & 5.18 & 0.64 & 0.60 & 0.53 & 4.48 & 0.82 & 0.33 \\
 & T+I & 0.66 & 7.61 & 0.97 & 0.98 & 0.56 & 6.28 & 0.96 & 0.94 \\
 & Tr & 0.53 & 4.31 & 0.48 & 0.66 & 0.49 & 3.64 & 0.62 & 0.49 \\
 & Tr+I & 0.62 & 6.87 & 0.84 & 0.94 & 0.52 & 5.38 & 0.81 & 0.89 \\
 & S & 0.59 & 4.94 & 0.43 & 0.76 & 0.48 & 2.79 & 0.51 & 0.63 \\
 & S+I & 0.64 & 7.22 & 0.93 & 0.92 & 0.49 & 4.50 & 0.72 & 0.74 \\
\midrule

Gemini-2.0-Flash & T & 0.55 & 5.49 & 0.70 & 0.73 & 0.53 & 5.71 & 0.60 & 0.77 \\
 & T+I & 0.61 & 6.90 & 0.93 & 0.98 & 0.55 & 6.69 & 0.94 & 0.98 \\
 & Tr & 0.52 & 4.65 & 0.54 & 0.75 & 0.50 & 4.83 & 0.42 & 0.80 \\
 & Tr+I & 0.59 & 6.29 & 0.75 & 0.96 & 0.53 & 6.18 & 0.79 & 0.93 \\
 & S & 0.46 & 3.23 & 0.30 & 0.86 & 0.46 & 2.87 & 0.37 & 0.79 \\
 & S+I & 0.61 & 6.37 & 0.93 & 0.97 & 0.53 & 5.92 & 0.90 & 0.92 \\
\midrule

Phi-4 & T & 0.56 & 5.10 & 0.46 & 0.57 & 0.51 & 3.17 & 0.66 & 0.47 \\
 & T+I & 0.57 & 6.12 & 0.80 & 0.95 & 0.49 & 3.75 & 0.81 & 0.81 \\
 & Tr & 0.53 & 4.31 & 0.42 & 0.61 & 0.48 & 2.72 & 0.53 & 0.57 \\
 & Tr+I & 0.55 & 5.52 & 0.68 & 0.93 & 0.46 & 3.22 & 0.66 & 0.77 \\
 & S & 0.35 & 1.58 & 0.32 & 0.72 & 0.31 & 1.25 & 0.57 & 0.43 \\
 & S+I & 0.52 & 4.36 & 0.62 & 0.69 & 0.39 & 2.14 & 0.51 & 0.50 \\
\midrule

Qwen-2.5-3B & T & 0.55 & 4.89 & 0.51 & 0.75 & 0.53 & 3.94 & 0.85 & 0.30 \\
 & T+I & 0.53 & 5.52 & 0.96 & 0.97 & 0.52 & 4.90 & 0.83 & 0.90 \\
 & Tr & 0.52 & 4.08 & 0.43 & 0.75 & 0.50 & 3.16 & 0.80 & 0.31 \\
 & Tr+I & 0.50 & 4.75 & 0.83 & 0.95 & 0.46 & 3.73 & 0.74 & 0.84 \\
 & S & 0.46 & 3.23 & 0.34 & 0.85 & 0.46 & 2.15 & 0.67 & 0.45 \\
 & S+I & 0.45 & 4.03 & 0.89 & 0.84 & 0.43 & 2.95 & 0.80 & 0.49 \\
\midrule

FT (Qwen-2.5-3B) & T & 0.73 & 6.52 & 0.91 & 0.88 & 0.65 & 5.93 & 0.90 & 0.85 \\
 & T+I & 0.78 & 8.50 & \textbf{0.98} & \textbf{0.99} & 0.68 & 7.60 & \textbf{0.97} & 0.98 \\
 & Tr & 0.69 & 5.64 & 0.86 & 0.82 & 0.62 & 5.19 & 0.85 & 0.77 \\
 & Tr+I & 0.76 & 8.01 & 0.91 & 0.95 & 0.66 & 7.23 & 0.90 & 0.93 \\
 & S & 0.71 & 6.10 & 0.90 & 0.82 & 0.62 & 4.97 & 0.82 & 0.76 \\
 & S+I & 0.77 & 8.37 & 0.95 & 0.94 & 0.66 & 7.22 & 0.87 & 0.86 \\
\bottomrule
\end{tabular}
}
\end{table*}

\begin{figure*}[t]
\centering
\begin{subfigure}{0.49\textwidth}
\centering
\begin{tikzpicture}
\begin{axis}[
ybar=2pt, bar width=6pt,
ymin=0, ymax=8.0,
width=\linewidth, height=6cm,
xtick=data,
xticklabels={T, T+I, Tr, Tr+I, A, A+I},
x tick label style={font=\footnotesize},
y tick label style={font=\footnotesize},
legend cell align=left,
legend style={at={(0.5,1.02)}, anchor=south, legend columns=-1, font=\footnotesize},
ylabel={Judge},
title={}
]
\addplot+[ybar, draw=white, fill=nxOrange] coordinates
{(1,5.18) (2,7.61) (3,4.31) (4,6.87) (5,4.94) (6,7.22)}; %
\addplot+[ybar, draw=white, fill=nxBlue] coordinates
{(1,5.49) (2,6.90) (3,4.65) (4,6.29) (5,3.23) (6,6.37)}; %
\legend{Qwen-7B, Gemini}
\end{axis}
\end{tikzpicture}
\end{subfigure}
\hfill
\begin{subfigure}{0.49\textwidth}
\centering
\begin{tikzpicture}
\begin{axis}[
ybar=2pt, bar width=6pt,
ymin=0, ymax=7.5,
width=\linewidth, height=6cm,
xtick=data,
xticklabels={T, T+I, Tr, Tr+I, A, A+I},
x tick label style={font=\footnotesize},
y tick label style={font=\footnotesize},
legend cell align=left,
legend style={at={(0.5,1.02)}, anchor=south, legend columns=-1, font=\footnotesize},
ylabel={Judge},
title={}
]
\addplot+[ybar, draw=white, fill=nxOrange] coordinates
{(1,4.48) (2,6.28) (3,3.64) (4,5.38) (5,2.79) (6,4.50)}; %
\addplot+[ybar, draw=white, fill=nxBlue] coordinates
{(1,5.71) (2,6.69) (3,4.83) (4,6.18) (5,2.87) (6,5.92)}; %
\legend{Qwen-7B, Gemini}
\end{axis}
\end{tikzpicture}
\end{subfigure}

\vspace{0.5em}

\begin{subfigure}{0.49\textwidth}
\centering
\begin{tikzpicture}
\begin{axis}[
ybar=2pt, bar width=6pt,
ymin=0, ymax=9.0,
width=\linewidth, height=6cm,
xtick=data,
xticklabels={T, T+I, Tr, Tr+I, S, S+I},
x tick label style={font=\footnotesize},
y tick label style={font=\footnotesize},
legend cell align=left,
legend style={at={(0.5,1.02)}, anchor=south, legend columns=-1, font=\footnotesize},
ylabel={Judge},
title={}
]
\addplot+[ybar, draw=white, fill=nxOrange] coordinates
{(1,4.89) (2,5.52) (3,4.08) (4,4.75) (5,3.23) (6,4.03)};
\addplot+[ybar, draw=white, fill=nxBlue] coordinates
{(1,6.52) (2,8.50) (3,5.64) (4,8.01) (5,6.10) (6,8.37)};
\legend{Qwen-3B, FT Qwen-3B}
\end{axis}
\end{tikzpicture}
\end{subfigure}
\hfill
\begin{subfigure}{0.49\textwidth}
\centering
\begin{tikzpicture}
\begin{axis}[
ybar=2pt, bar width=6pt,
ymin=0, ymax=8.0,
width=\linewidth, height=6cm,
xtick=data,
xticklabels={T, T+I, Tr, Tr+I, S, S+I},
x tick label style={font=\footnotesize},
y tick label style={font=\footnotesize},
legend cell align=left,
legend style={at={(0.5,1.02)}, anchor=south, legend columns=-1, font=\footnotesize},
ylabel={Judge},
title={}
]
\addplot+[ybar, draw=white, fill=nxOrange] coordinates
{(1,3.94) (2,4.90) (3,3.16) (4,3.73) (5,2.15) (6,2.95)};
\addplot+[ybar, draw=white, fill=nxBlue] coordinates
{(1,5.93) (2,7.60) (3,5.19) (4,7.23) (5,4.97) (6,7.22)};
\legend{Qwen-3B, FT Qwen-3B}
\end{axis}
\end{tikzpicture}
\end{subfigure}

\caption{LLM-Judge scores across modalities. Top: Qwen2.5-7B (Qwen-7B) vs Gemini 2.5-pro (Gemini) for English (left) and MSA (right). Bottom: Qwen2.5-3B (Qwen-3B) vs its fine-tuned variant for English (left) and MSA (right).}
\label{fig:judge_merged}
\end{figure*}

\textbf{Image Modality Improves Overall Performance} 
Based on our analysis, we observe that T+I performance for Arabic varieties is relatively higher than for English.  Table~\ref{tab:oe_gain_all_models} summarizes the average T$\rightarrow$T+I improvement for open-ended QA across the four language variants. Adding images consistently improves LLM-as-judge scores for all models, with the largest gains for GPT-4.1 (+2.14) and GPT-5 (+1.80), followed by the fine-tuned Qwen model (+1.66). F1 gains are smaller overall, averaging +0.026 across models, indicating that image context improves answer quality more strongly under semantic judgment than token-level similarity. The only negative mean F1 change appears for Qwen-3B, although its judge score still improves by +0.81.

\begin{table}[t]
\centering
\caption{Mean absolute T$\rightarrow$T+I gains for open-ended QA across language variants.}
\label{tab:oe_gain_all_models}
\setlength{\tabcolsep}{6pt}
\scalebox{0.90}{
\begin{tabular}{lcc}
\toprule
\textbf{Model} & \textbf{OE (F1)} & \textbf{OE (Judge)} \\
\midrule
GPT-4.1 & +0.070 $\pm$ 0.041 & +2.138 $\pm$ 0.176 \\
GPT-5 & +0.028 $\pm$ 0.015 & +1.795 $\pm$ 0.184 \\
Gemini & +0.030 $\pm$ 0.022 & +1.325 $\pm$ 0.211 \\
Qwen-7B & +0.030 $\pm$ 0.027 & +0.788 $\pm$ 0.916 \\
Phi-4 & +0.013 $\pm$ 0.019 & +0.700 $\pm$ 0.347 \\
Qwen-3B & -0.023 $\pm$ 0.017 & +0.813 $\pm$ 0.261 \\
FT (Qwen-2.5-3B) & +0.035 $\pm$ 0.010 & +1.655 $\pm$ 0.172 \\
\midrule
\textbf{Overall} & \textbf{+0.026 $\pm$ 0.033} & \textbf{+1.316 $\pm$ 0.645} \\
\bottomrule
\end{tabular}
}
\end{table}

\section{Generalization Study} 
In Table~\ref{tab:results_alm_bench}, we provide additional analysis of the ALM-Bench dataset by reporting performance \emph{by question type} to better assess our model's generalization. Specifically, we evaluate MCQ, true/false (T/F), and open-ended (OE) questions (both short and long), and we compare results for Arabic and its English translation. The overall LLM-as-a-judge score for Arabic increases from 4.6 for the base model to 5.8 for our fine-tuned model, indicating improved capability after training on \oasis{}. We observe a similar trend on the English translation of the Arabic subset, where the fine-tuned model also achieves higher overall scores. Overall, performance on the English questions is higher than on the Arabic questions.

\begin{table}[h]
\centering
\caption{Results (Acc) on the Arabic subset of ALM-bench using Qwen-3B in its base and fine-tuned variant, along with the Qwen-VL model reported in \cite{vayani2025all}}
\label{tab:results_alm_bench_acc_llm_judge}
\setlength{\tabcolsep}{4pt} 
\scalebox{0.70}{
\begin{tabular}{@{}lrrrr@{}}
\toprule
\multicolumn{1}{c}{\textbf{Model}} & \multicolumn{1}{c}{\textbf{Saudi}} & \multicolumn{1}{c}{\textbf{Egyptian}} & \multicolumn{1}{c}{\textbf{Emirati}} & \multicolumn{1}{c}{\textbf{Avg}} \\ \midrule
Qwen-2-VL~\cite{vayani2025all}   & 0.59 & 0.62 & 0.57 & 0.59 \\
Qwen 3B     & 0.62 & 0.63 & 0.60 & 0.62 \\
Qwen 3B-FT  & \textbf{0.69} & \textbf{0.72} & \textbf{0.70} & \textbf{0.70} \\ \bottomrule
\end{tabular}
}
\vspace{-0.1cm}
\end{table}

\begin{table}[h]
\centering
\caption{Results on the Arabic subset of ALM-bench using Qwen-3B in its base and fine-tuned models. The numbers of questions for MCQ, T/F, and OE are 194, 99, and 449, respectively.}
\label{tab:results_alm_bench}
\setlength{\tabcolsep}{2pt} 
\scalebox{0.65}{
\begin{tabular}{@{}llrrr@{}}
\toprule
\textbf{Lang.} & \textbf{Model} & \multicolumn{1}{l}{\textbf{MCQ (Acc)}} & \multicolumn{1}{l}{\textbf{TF (Acc)}} & \multicolumn{1}{l}{\textbf{OE (Judge)}} \\ \midrule
Ar & Qwen-3B & 76.8 & 90.9 & 4.6 \\
Ar & FT (Qwen-3B) & \textbf{86.6} & 89.9 & \textbf{5.8} \\ \midrule
En & Qwen-3B & 83.6 & 85.8 & 5.7 \\
En & FT (Qwen-3B) & \textbf{92.8} & \textbf{90.9} & \textbf{7.1} \\ \bottomrule
\end{tabular}
}
\vspace{-0.2cm}
\end{table}

\section{Qualitative Analysis}
\label{sec_app_aualitative_analysis}

\subsection{Error Analysis}
\begin{figure*}[tbh!]
    \centering
    \includegraphics[width=0.95\linewidth]{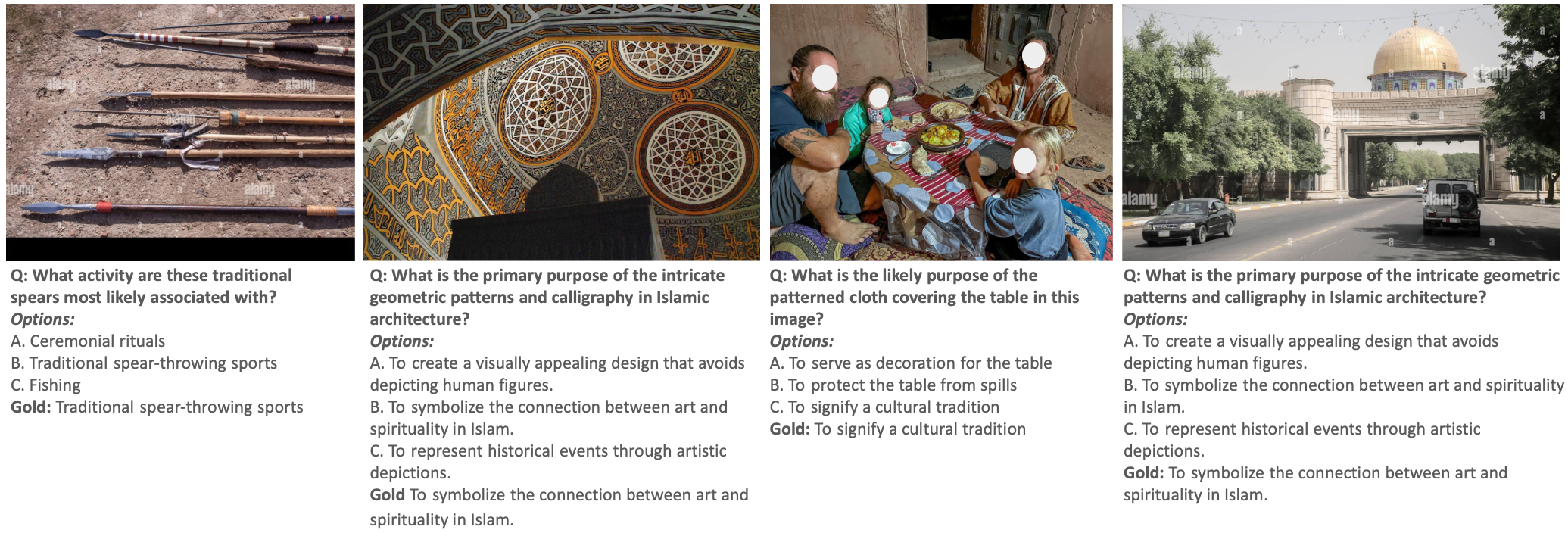}
    \caption{Examples of images where models incorrectly answered.}
    \label{fig:examples_images_error}
\end{figure*}

We conduct an error analysis to identify the image and question types that models fail to answer correctly. Figure~\ref{fig:examples_images_error} shows examples where almost all models fail. These cases indicate that failures are often not due to visual recognition alone, but also to missing contextual, cultural, or locale-specific knowledge.
We further analyze recurring error patterns. The main failure modes include culturally ambiguous visual cases that lead to incorrect cultural attribution, and translation artifacts such as lexical transfer and dialect mixing. For Algerian open-ended QA in MSA, culturally grounded interpretation questions constitute 26.6\% of the data, on which 68.2\% shows errors, with substantially lower scores than attribute-recognition questions (5.21 vs.\ 7.49). The dominant errors are under-specified answers and culturally incorrect hallucinations. 

\subsection{Category-wise Analysis} 
\begin{figure*}[tbh!]
    \centering
    \begin{subfigure}[b]{0.48\linewidth}
        \centering
        \includegraphics[width=\linewidth]{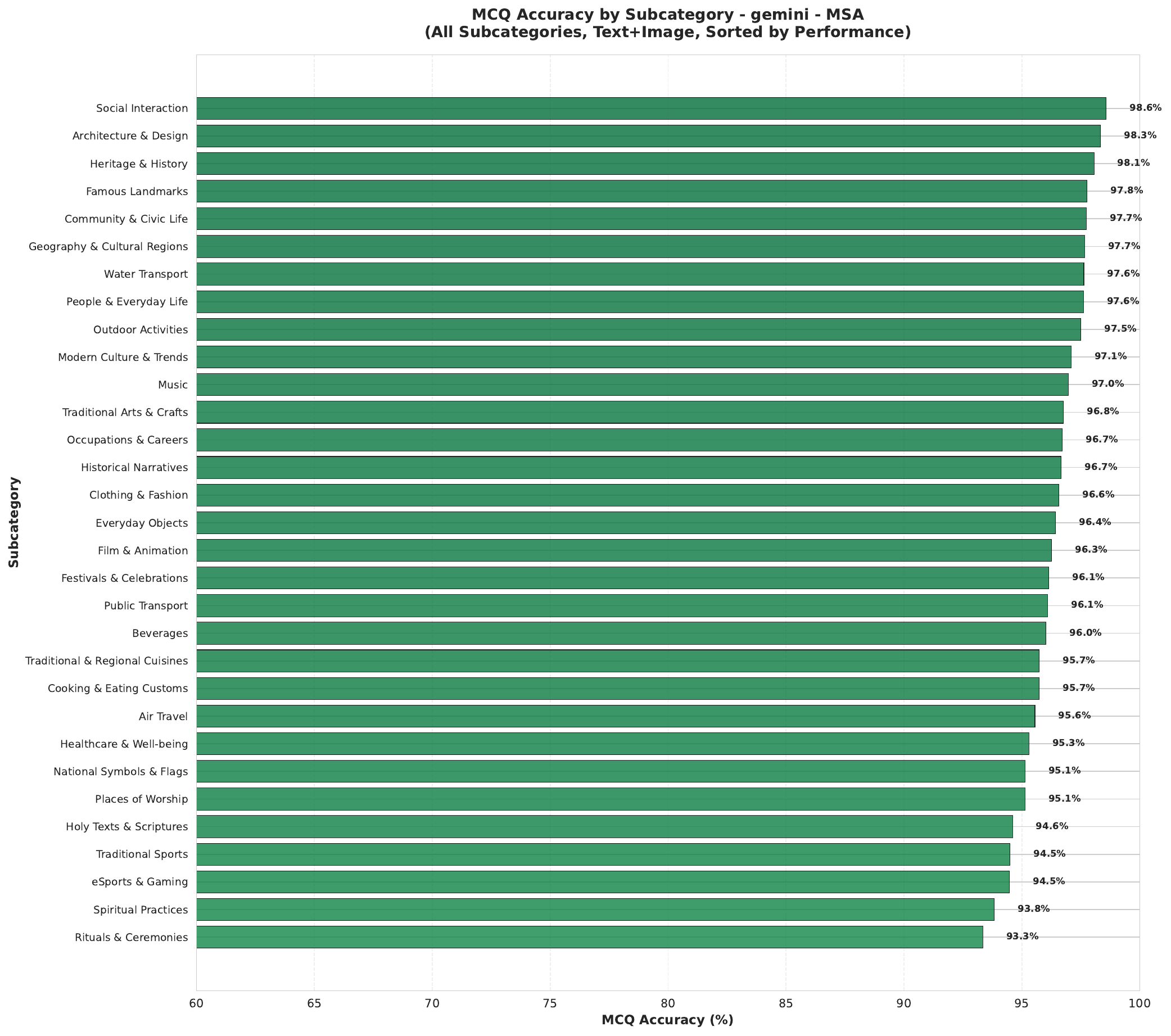}
        \caption{MSA}
        \label{fig:subcategory_mcq_gemini_msa}
    \end{subfigure}
    \hfill
    \begin{subfigure}[b]{0.48\linewidth}
        \centering
        \includegraphics[width=\linewidth]{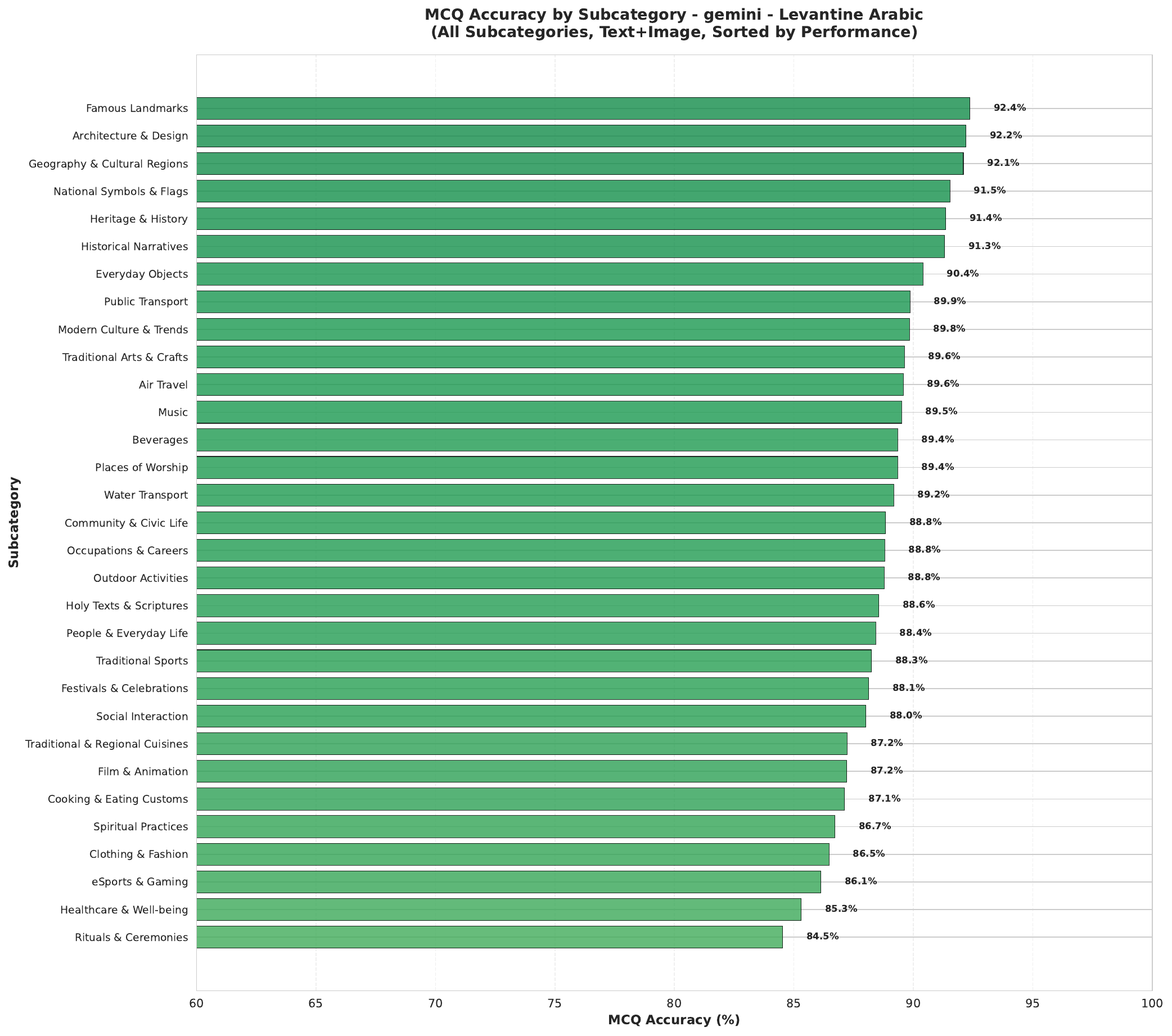} 
        \caption{Levantine}
        \label{fig:subcategory_mcq_gemini_2}
    \end{subfigure}
    \begin{subfigure}[b]{0.48\linewidth}
        \centering
        \includegraphics[width=\linewidth]{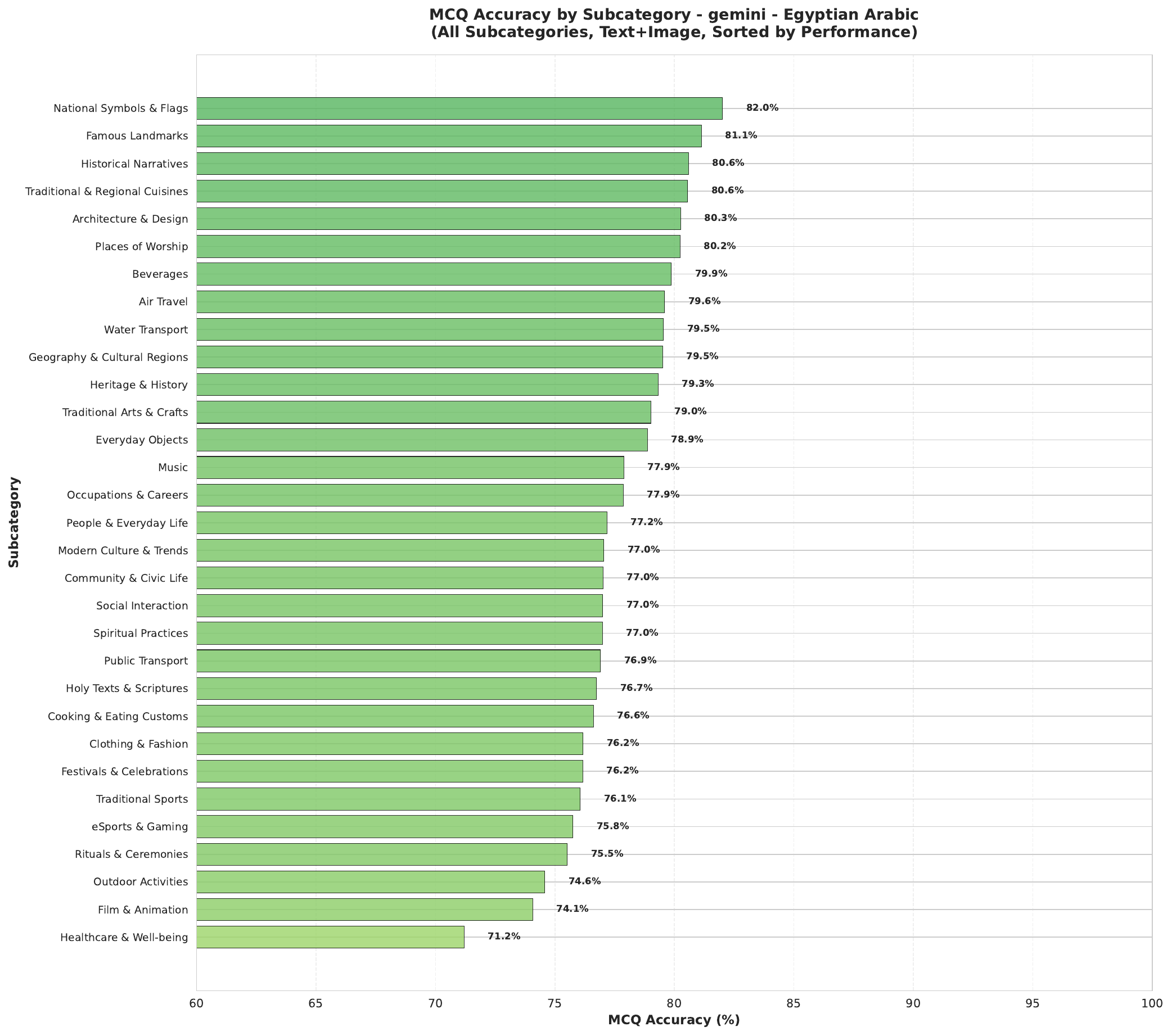}
        \caption{Egpytian}
        \label{fig:subcategory_mcq_gemini_3}
    \end{subfigure}
    \caption{Subcategory-wise MCQ performance across three Arabic variants using Gemini.}
    \label{fig:subcategory_mcq_gemini_all}
\end{figure*}

\begin{figure*}
    \centering
    \includegraphics[width=0.85\linewidth]{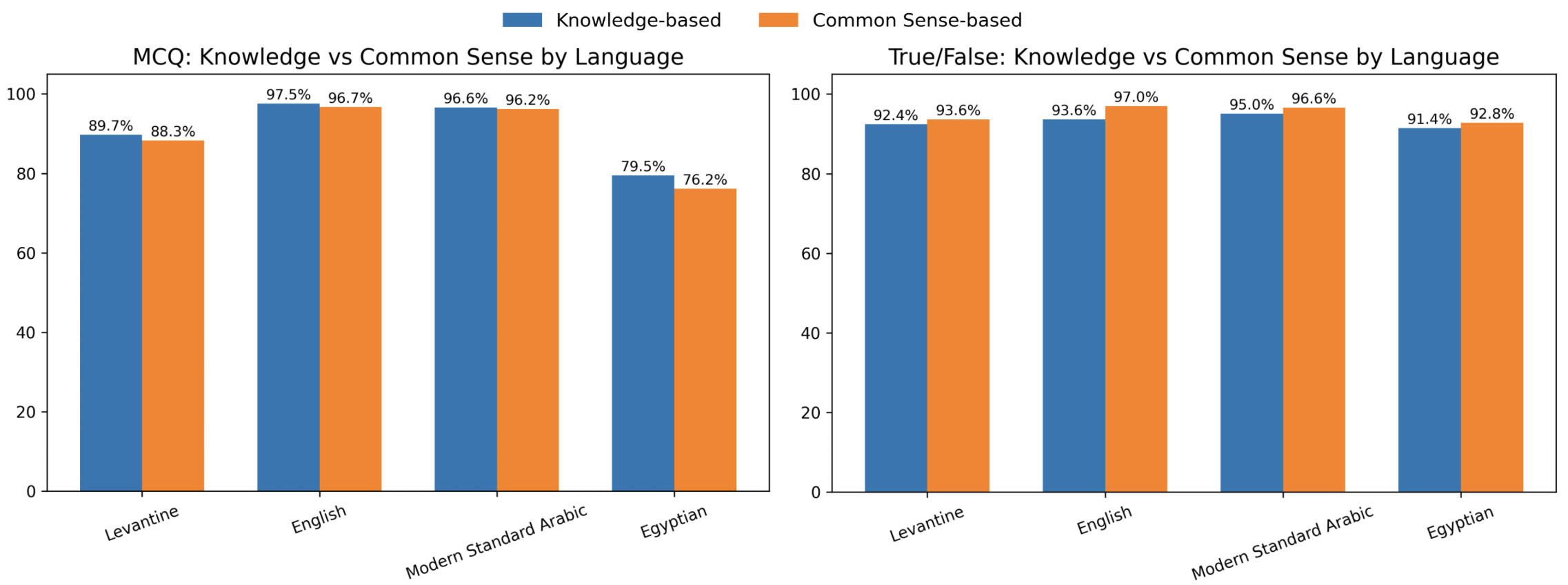}
    \caption{Comparison of \textit{knowledge} vs. \textit{commonsense} MCQ performance of the Gemini model across different language variants.}
    \label{fig:knowledge_vs_commonsense}
\end{figure*}

\begin{figure}[t]
    \centering
    \begin{subfigure}[b]{1.0\linewidth}
        \centering
        \includegraphics[width=\linewidth]{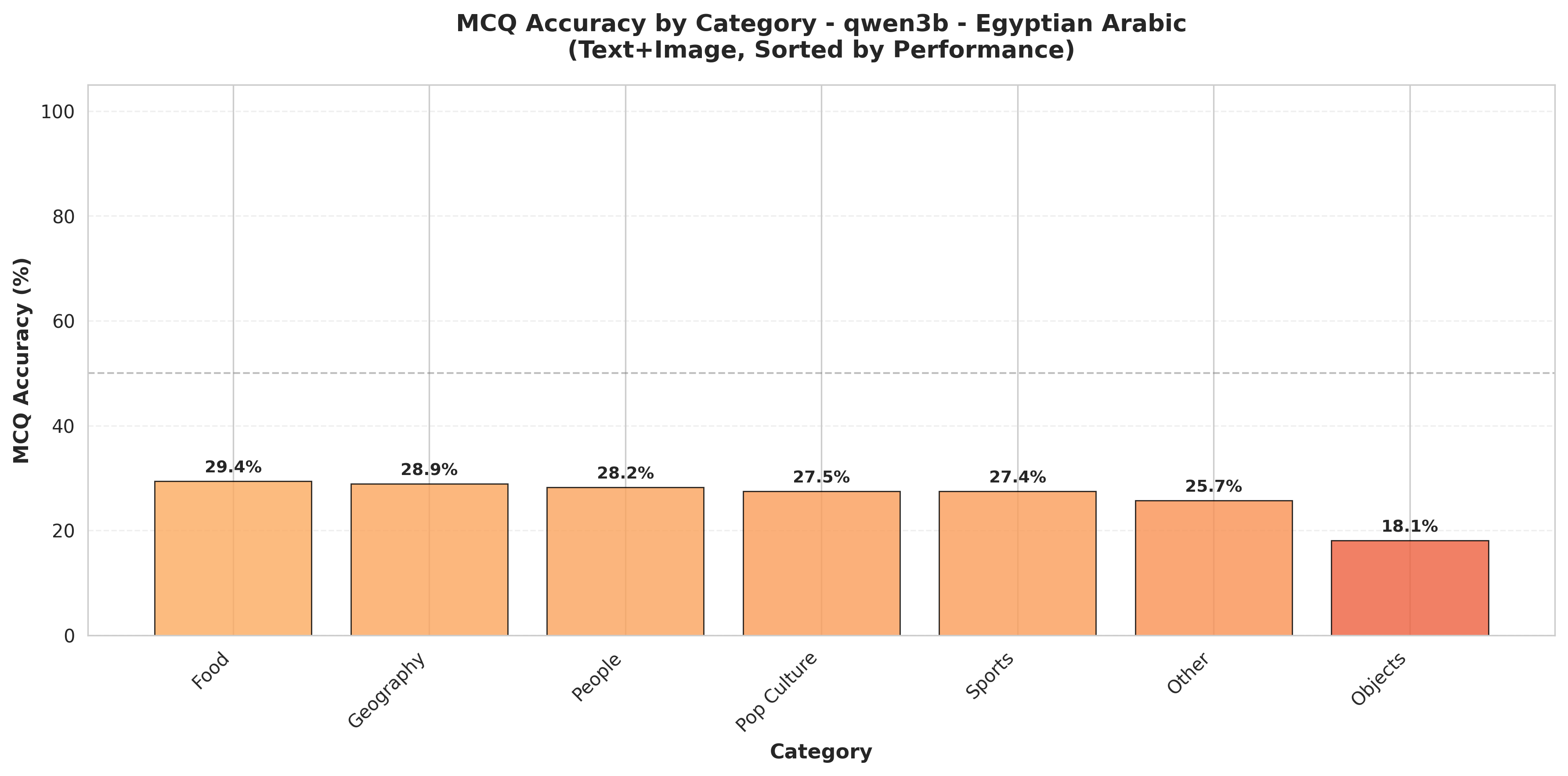}
        \caption{Qwen3b base model.}
        \label{fig:subcategory_mcq_qwen3b_1}
    \end{subfigure}
    
    \vspace{0.6em}
    
    \begin{subfigure}[b]{1.0\linewidth}
        \centering
        \includegraphics[width=\linewidth]{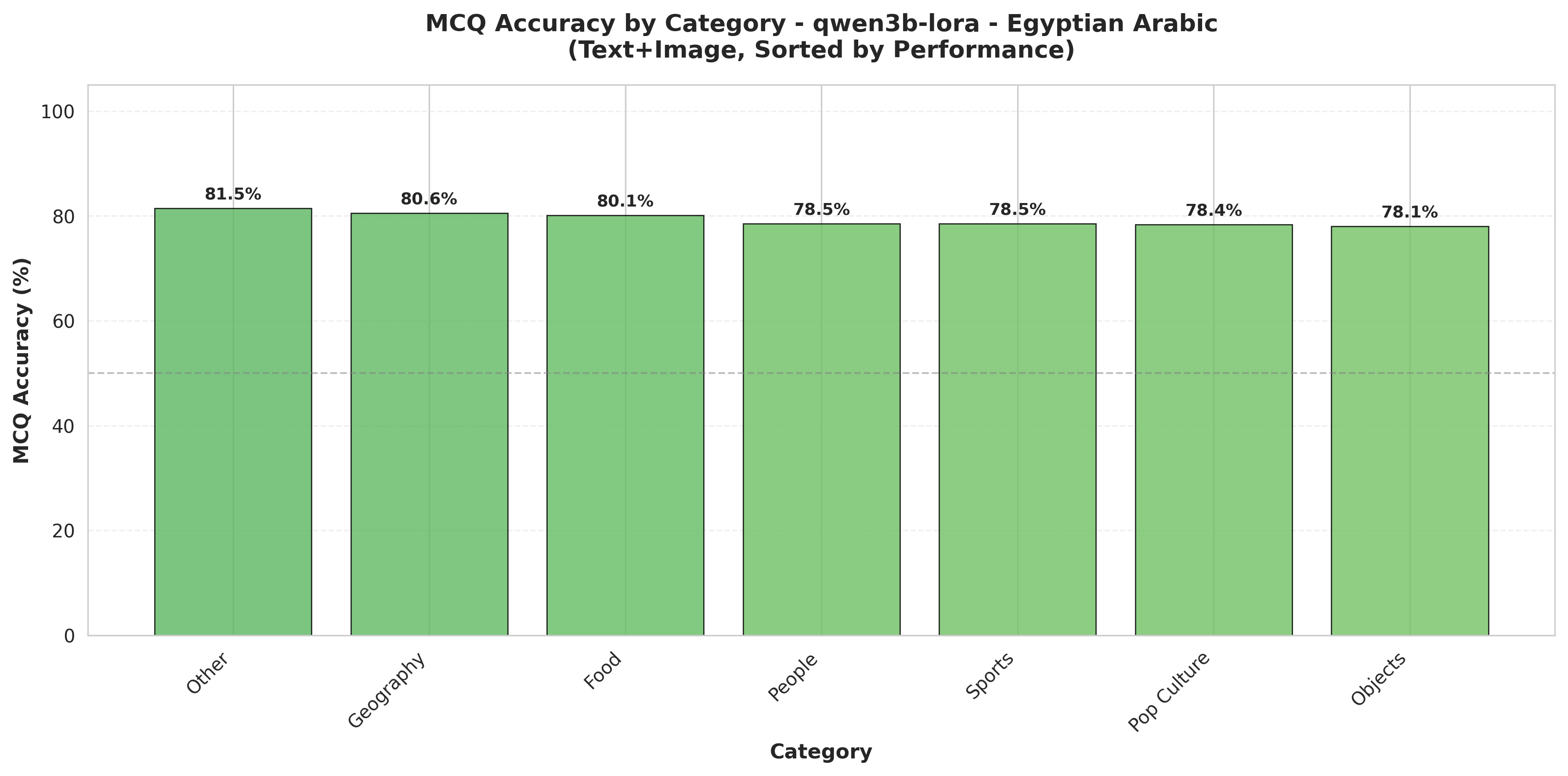} 
        \caption{Qwen3b fine-tuned model.}
        \label{fig:subcategory_qwen3b_2}
    \end{subfigure}
    \caption{Category-wise MCQ performance on Egyptian dialects with and without the fine-tuned Qwen-3B model.}
    \label{fig:subcategory_mcq_qwen3b}
\end{figure}

We also performed category- and subcategory-wise performance analysis across all models. Our findings show that models perform well in some categories (e.g., Heritage \& History) but struggle in others (e.g., eSports \& Gaming), as presented in Figure \ref{fig:subcategory_mcq_gemini_all}. In Figure \ref{fig:subcategory_mcq_qwen3b}, we show category-wise MCQ performance for Egyptian dialects using the Qwen-3B base and fine-tuned models. The results demonstrate that fine-tuning significantly improves performance in several categories (e.g., Objects).
In Figure \ref{fig:knowledge_vs_commonsense}, we report the performance by grouping knowledge vs. commonsense based QA, which are obtained using the Gemini model. 

\begin{figure*}[tbh!]
    \centering
    \includegraphics[width=0.75\linewidth]{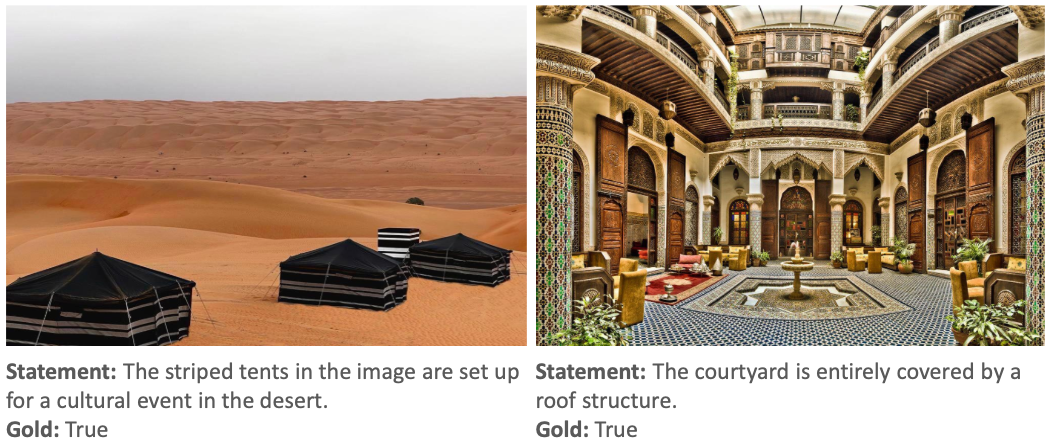}
    \caption{Examples of images where the fine-tuned model answered correctly.}
    \label{fig:examples_images_ft_correct}
\end{figure*}

\section{Related Work -- Additional Details}
Table~\ref{tab:related-work} provides a comparative overview of existing multimodal and multilingual benchmarks. It summarizes each benchmark’s supported modalities (text, image, speech), multilingual coverage, number of language varieties and scripts, domains, dataset size, question types and forms, and annotation methods. The table highlights differences in scale, linguistic diversity, and task design across benchmarks, illustrating where \oasis{} fits in terms of multimodality, multilingual support, dataset size, and question diversity.

\begin{table*}[t]
\centering
\caption{Comparison of multimodal and multilingual benchmarks. 
\textbf{Mod}: Modalities (Text = T, Image = I, Speech = S). 
\textbf{Multi}: Multilingual support. 
\textbf{Lang}: \# of languages varieties. 
\textbf{Script}: \# of writing scripts. 
\textbf{Dom}: \# of domains. 
\textbf{Samp}: Total samples. 
\textbf{QTypes}: Question types (MCQ = multiple-choice, SVQA = short visual QA, LVQA = long visual QA, TF = true/false, OE = open-ended, Y/N = yes/no). 
\textbf{QForms}: Question forms (Fixed or Diverse). 
\textbf{Annot}: Annotation type (Auto = automatic, Manual = human, Auto+Manual = hybrid). $^{*}$ 10K QA pairs associated with 5,239 images. $^{\dagger}$ 1,999 QA pairs associated with 515 images.}
\label{tab:related-work}
\setlength{\tabcolsep}{3pt}
\scalebox{0.7}{
\begin{tabular}{lcccccccccc}
\toprule
\textbf{Benchmark} & \textbf{Mod} & \textbf{Multi} & \textbf{Lang.} & \textbf{Script} & \textbf{Dom} & \textbf{Samp} & \textbf{QTypes} & \textbf{QForms} & \textbf{Annot} \\
\midrule
CVQA \citep{romero2025cvqa}   & T,I   & \checkmark & 31 & 13 & 10 & 5,239$^{*}$  & MCQ & Fixed & Manual \\
ALM-Bench \citep{vayani2025all}   & T,I   & \checkmark & 100 & 24 & 19 & 22,763  & MCQ, SVQA, LVQA, TF & Diverse & Auto+Manual \\
CulturalVQA \citep{nayak-etal-2024-benchmarking}   & T,I   & \checkmark & 1 & 1 & 5 & 2,378  & SVQA & - & Manual \\
SeaVQA \citep{urailertprasert-etal-2024-sea}   & T,I   & \checkmark & 1 & 1 & - & 515$^{\dagger}$ & MCQ & - & Manual \\
Camel-Bench \citep{ghaboura-etal-2025-camel} & T,I & \checkmark & 2 & 2 & 5 & 29,036 & SVQA, LVQA & Diverse & Auto+Manual \\
MM-Vet \citep{yu2024mm}            & T,I   & $\times$ & 1 & 1 & 16 & 218 & SVQA, LVQA & Fixed & Manual \\
Pangea-Bench \citep{yue2024pangea} & T,I   & \checkmark & 47 & 13 & 18 & - & MCQ, SVQA & Fixed & Auto \\
MMBench \citep{liu2024mmbench}     & T,I   & \checkmark & 2 & 2 & 20 & 3,217 & MCQ & Fixed & Manual \\
MaRVL \citep{collini2025marvel}    & T,I   & \checkmark & 5 & 3 & 11 & 5,670 & TF & Fixed & Manual \\
M3Exam \citep{zhang2023m3exam}     & T,I   & \checkmark & 9 & 3 & 4 & 12,317 & MCQ & Diverse & - \\
xGQA \citep{pfeiffer2022xgqa}      & T,I   & \checkmark & 8 & 5 & - & 12,578 & Y/N, SVQA & Fixed & - \\
OmniBench \citep{li2024omnibench}  & T,I,S & \checkmark & 2 & 2 & 8 & 1,142 & MCQ & Diverse & Manual \\
Pearl \citep{alwajih2025pearl}     & T,I   & $\times$ & 1 & 1 & 10 & 309,000 & 13 types & Diverse & Auto+Manual \\ \midrule
\oasis{}                           & T,I,S & \checkmark & 2 & 4 & 31 & 0.92M & OE, MCQ, TF & Diverse & Auto+Manual \\
\bottomrule
\end{tabular}
}
\end{table*}

\section{Dataset Sample Examples}
\label{sec_app_dataset_examples}
\begin{figure*}
    \centering
    \includegraphics[width=0.95\linewidth]{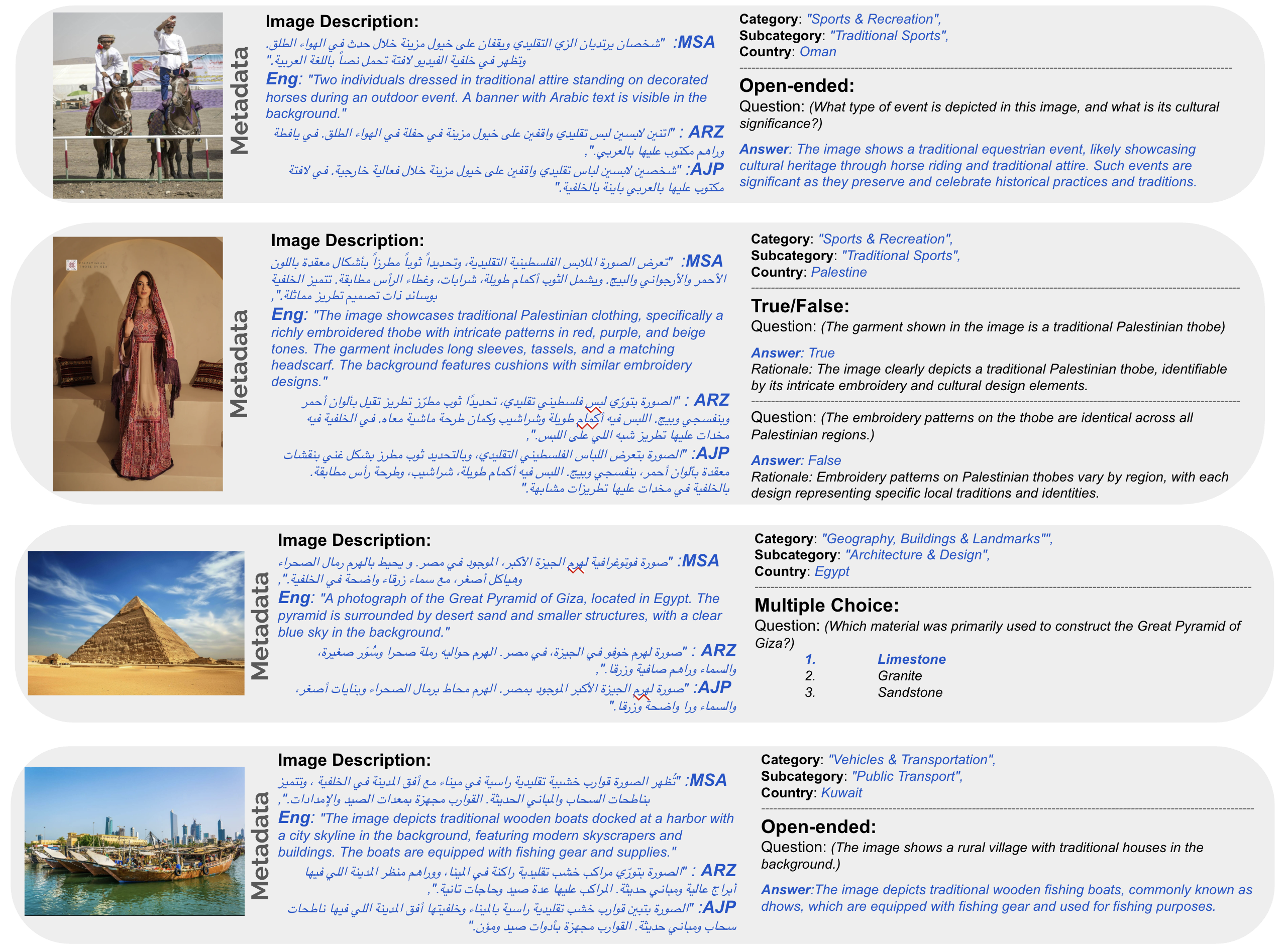}
    \caption{Examples.}
    \label{fig:example_set2}
\end{figure*}

\begin{figure*}
    \centering
    \includegraphics[width=0.95\linewidth]{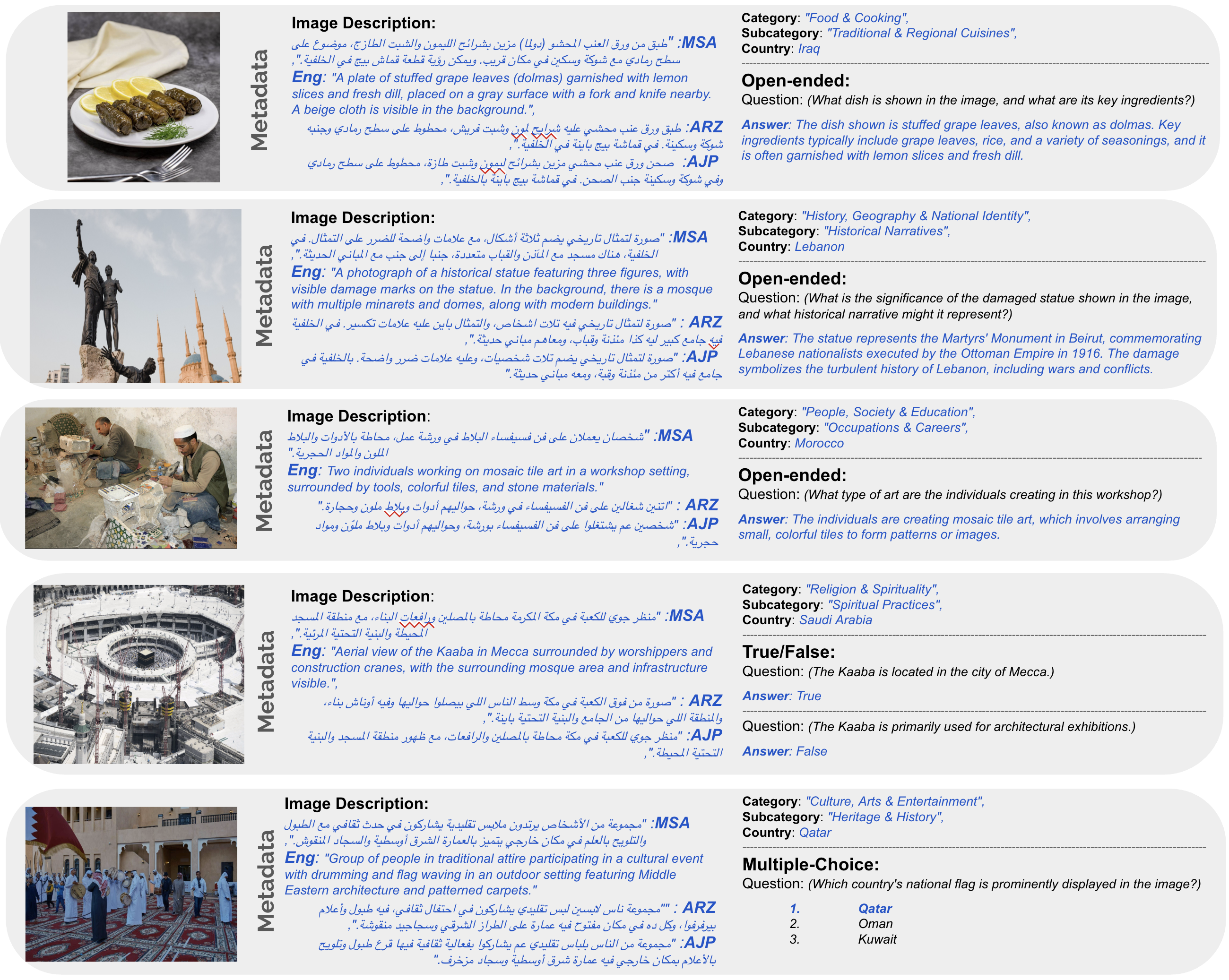}
    \caption{Examples.}
    \label{fig:example_set1}
\end{figure*}

In this section, we provide example images, as shown in Figures \ref{fig:example_set2} and \ref{fig:example_set1}, along with their associated metadata, language variants, and corresponding open-ended questions. 

\section{Annotation Guidelines}
\label{app_section_annotation_guidelines}

This section provides the annotation guidelines used for \textit{(i)} voice recording and \textit{(ii)} quality assessment of QA pairs. These instructions were shown to annotators during data collection to ensure consistency, clarity, and high-quality annotations across countries and languages.

\subsection{QA Annotation Guidelines}
\label{appendix:annotation_guidelines}
Annotators evaluated image-based question–answer (QA) pairs including one open-ended question, one multiple-choice question, and two true/false items.
On the annotation interface, you will see the following:
\begin{enumerate}
    \item An image
    \item A description of the image to help you understand the image better
    \item The different types of questions, answers, and a rationale for each answer, all related to the image shown
\end{enumerate}
The types of questions will be:
\begin{itemize}
    \item One open-ended question
    \item One multiple-choice question with choices
    \item Two true/false questions with the selected answer
\end{itemize}
\subsubsection{Annotation Task}
\begin{itemize}
    \item Decide if the image and its associated questions are related to the specified location.
    \item For each type of question, score the clarity and quality on a scale from 1--5.
    \item Indicate whether answering the question requires external knowledge (e.g., searching online or consulting information not present in the image).
    \item For each answer, score the correctness (1--5 for open-ended and multiple-choice; 1--3 for true/false).
    \item For each rationale, score the quality: Clarity, Informativeness, Plausibility, Faithfulness (1--5).
    \item If any score is less than 4, choose a reason for revision.
\end{itemize}

\subsubsection{Scoring Questions and Answers}

\noindent
\textbf{Open-Ended Question and its Answer}

\textbf{Question Quality (1--5):} Assess clarity, relevance, and lack of ambiguity.  
Revision reasons (if score < 4):
\begin{itemize}[noitemsep,topsep=0pt,leftmargin=*,labelsep=.5em]
    \item Unclear or ambiguous
    \item Not relevant to the image
    \item Hard to understand
    \item Requires external knowledge
\end{itemize}

\textbf{Answer Quality (1--5):} Assess factuality, conciseness, and grounding in the image.  
Revision reasons:
\begin{itemize}[noitemsep,topsep=0pt,leftmargin=*,labelsep=.5em]
    \item Incorrect or unsupported by the image
    \item Incomplete or missing key information
    \item Speculative or assumption-based
\end{itemize}

\textbf{Rationale Quality:}  
\begin{itemize}[noitemsep,topsep=0pt,leftmargin=*,labelsep=.5em]
    \item Clarity \& Informativeness (1--5)
    \item Plausibility \& Faithfulness (1--5)
\end{itemize}

\noindent
\textbf{Multiple-Choice Question}

\textbf{Question Quality (1--5):} Clarity, specificity, relevance.  
Revision reasons (if score < 4):
\begin{itemize}[noitemsep,topsep=0pt,leftmargin=*,labelsep=.5em]
    \item Unclear or ambiguous
    \item Not relevant to the image
    \item Hard to understand
\end{itemize}

\textbf{Answer Quality (1--5):} Correctness and image support.  
Revision reasons:
\begin{itemize}[noitemsep,topsep=0pt,leftmargin=*,labelsep=.5em]
    \item Options overlap in meaning
    \item Irrelevant or implausible options
    \item Vague or confusing options
\end{itemize}

\textbf{Rationale Quality:}  
\begin{itemize}[noitemsep,topsep=0pt,leftmargin=*,labelsep=.5em]
    \item Clarity \& Informativeness (1--5)
    \item Plausibility \& Faithfulness (1--5)
\end{itemize}

\noindent
\textbf{True/False Questions and Selected Answer}

\textbf{Statement Quality (1--5):} Clarity, factual nature, verifiability from the image.  
Revision reasons (if score < 4):
\begin{itemize}[noitemsep,topsep=0pt,leftmargin=*,labelsep=.5em]
    \item Unclear or ambiguous
    \item Not factual
\end{itemize}

\textbf{Selected Answer Quality (1--3):} Correctness and factual grounding.  
Revision reasons:
\begin{itemize}[noitemsep,topsep=0pt,leftmargin=*,labelsep=.5em]
    \item Incorrect or unsupported by the image
    \item Incomplete or missing information
    \item Speculative or assumption-based
\end{itemize}

\textbf{Rationale Quality:}  
\begin{itemize}[noitemsep,topsep=0pt,leftmargin=*,labelsep=.5em]
    \item Clarity \& Informativeness (1--5)
    \item Plausibility \& Faithfulness (1--5)
\end{itemize}

\noindent
\textbf{General Annotation Principles}
\begin{itemize}[noitemsep,topsep=0pt,leftmargin=*,labelsep=.5em]
    \item Avoid speculation beyond what is visible or inferable from the image.
    \item Mark questions requiring external knowledge.
    \item Always select a revision reason for scores below 4.
    \item You may enlarge the image by opening it in a new window.
\end{itemize}

\subsection{Voice Recording Instructions}
\label{appendix:recording}
Annotators were asked to record spoken the question %
The following guidelines were displayed in the interface:
\begin{itemize}[noitemsep,topsep=0pt,leftmargin=*,labelsep=.5em]
    \item \textbf{Read the Sentence:} A sentence appears on the screen. Read it aloud clearly.
    \item \textbf{Record:} Click \emph{Record} to start capturing your voice. Speak naturally and clearly.
    \item \textbf{Playback:} After finishing, click \emph{Play} to listen to your recording.
    \item \textbf{Review:} If satisfied with the audio, click \emph{Submit} to save it.
    \item \textbf{Re-record if Needed:} If the recording is unclear or incorrect, click \emph{Delete} and 
    rerecord the sentence.
    \item \textbf{Submit:} Once satisfied, click \emph{Submit} to store the final version and proceed.
\end{itemize}

\section{Limitations}
We designed \oasis{} and the \emqa{} framework to be practical and broadly reusable, while making several deliberate trade-offs.

\textit{First}, \emqa{} uses LLMs to support key steps, such as proposing candidate queries/questions and producing preliminary relevance scores. We mitigate potential bias through multi-stage human-in-the-loop validation, including filtering, manual QA checks, and human evaluation by annotators from multiple countries.

\textit{Second}, to separate scalable training from real-world evaluation, we use \emph{semi-synthetic speech based on neural voice cloning only for training} and report results on a \emph{human-recorded test set} covering diverse speakers and recording conditions. We also evaluate robustness by adding noise to recorded audio. While these tests provide a strong real-world benchmark, additional targeted evaluations, such as uncommon devices/channels and rare acoustic phenomena, would better characterize robustness under long-tail conditions.

\textit{Third}, due to computational constraints, we could not train models on the full dataset. With larger compute resources, future work can leverage the complete dataset to better explore its potential and assess the benefits of large-scale multimodal training.

\textit{Finally}, as with many multimodal QA benchmarks, some \oasis{} items may retain language priors. We view these as clear directions for iterative improvement.

\section{Broader Impact} \label{sec: broader_impact}
\oasis{} and \emqa{} aim to broaden participation in omnimodal research area by enabling the creation and evaluation of culturally grounded, location-specific, everyday information-seeking multimodal QA resources beyond English-centric settings. Our framework is modular and reusable, allowing researchers and practitioners to adapt the pipeline to new languages, regions, and domains. By lowering the barrier to building large-scale, culturally relevant benchmarks, we expect this work to accelerate progress in multilingual and dialectal multimodal understanding, support more equitable evaluation of foundation models, and promote the development of assistant technologies that work reliably for diverse user populations.

For downstream applications, \oasis{} supports the development of omnimodal models that better reflect everyday needs, such as tourism assistance, accessibility tools, education, and customer support. In addition, our evaluation setup includes a large human-recorded test set spanning diverse speakers and speaking conditions, which encourages robust model development.

Overall, we expect \emqa{} and \oasis{} to help shift omnimodal research toward culturally inclusive, deployment-relevant evaluation and model development.

\section{Ethics Statement} \label{sec: ethics}
We collected images in accordance with public-use licensing. Because image collection was driven by geo-location and query availability, some locations have fewer images (e.g., Qatar). For the manual recordings and annotations, contributors (42 speakers for recording and 52 annotators for QA checking) were compensated at standard hourly rates and were provided with task details in advance. As part of the recruitment process, they signed NDAs that permit us to use and disseminate the dataset. Overall, we do not foresee any ethical concerns arising from \oasis{}.

\section{Reproducibility statement} \label{sec:reproducability}
We made every effort to ensure reproducibility. The main paper details the \emqa{} framework and the construction of \oasis{}, along with the training and evaluation setups. Appendices~\ref{sec-app-prompt-topic}, \ref{sec-prompt-img-desc-gen}, and~\ref{sec-prompt-qa-gen} provide the prompts used to build \oasis{}, while Appendix~\ref{sec-ablation-topic-query} outlines query preprocessing and ablation configurations. 

\section{Data and Resource Release}
\label{sec:data_release}
We will release\footnote{\url{anonymous.com}} the supporting code, documentation, and dataset under the CC BY-NC-SA 4.0 license to facilitate adoption, reproducibility, and future extensions.

\end{document}